\documentclass[sigconf]{acmart} 
\renewcommand\footnotetextcopyrightpermission[1]{} 

\usepackage{booktabs}
\usepackage{multirow}
\usepackage{subcaption}
\usepackage{enumitem}

\def\imagebox#1#2{\vtop to #1{\null\hbox{#2}\vfill}}

\AtBeginDocument{%
  \providecommand\BibTeX{{%
    \normalfont B\kern-0.5em{\scshape i\kern-0.25em b}\kern-0.8em\TeX}}}

\setcopyright{acmcopyright}
\copyrightyear{2022}
\acmYear{2022}
\acmDOI{10.1145/1122445.1122456}

\acmConference[None]{In Proceedings of the Web Conference 2022 (WWW '22)}{--}{Online}
\acmBooktitle{In Proceedings of the Web Conference 2022 (WWW '22),
  April 25--29, 2022, Lyon, France}
\acmPrice{15.00}
\acmISBN{978-1-4503-XXXX-X/18/06}

\begin{document}

\title{CDGNet: A Cross-Time Dynamic Graph-based Deep Learning Model for Traffic Forecasting}

\author{Yuchen Fang}
 \affiliation{
   \institution{Beijing University of Posts and Telecommunications}
   \country{}
 }
\email{fangyuchen@bupt.edu.cn}

\author{Yanjun Qin}
 \affiliation{
   \institution{Beijing University of Posts and Telecommunications}
   \country{}
 }
\email{qinyanjun@bupt.edu.cn}

\author{Haiyong Luo}
\authornote{Corresponding author.}
 \affiliation{
   \institution{Institute of Computing Technology, Chinese Academy of Sciences}
   \country{}
 }
\email{yhluo@ict.ac.cn}

\author{Fang Zhao}
\authornotemark[1]
 \affiliation{
   \institution{Beijing University of Posts and Telecommunications}
   \country{}
 }
\email{zfsse@bupt.edu.cn}

\author{Liang Zeng}
 \affiliation{
   \institution{Tsinghua University}
   \country{}
 }
\email{zengl18@mails.tsinghua.edu.cn}

\author{Bo Hui}
 \affiliation{
   \institution{Auburn University}
   \country{}
 }
\email{bohui@auburn.edu}

\author{Chenxing Wang}
 \affiliation{
   \institution{Beijing University of Posts and Telecommunications}
   \country{}
 }
\email{wangchenxing@bupt.edu.cn}
\renewcommand{\shortauthors}{Fang, et al.}

\begin{abstract}
Traffic forecasting is important in intelligent transportation systems of webs and beneficial to traffic safety, yet is very challenging because of the complex and dynamic spatio-temporal dependencies in real-world traffic systems. Prior methods use the pre-defined or learnable static graph to extract spatial correlations. However, the static graph-based methods fail to mine the evolution of the traffic network. Researchers subsequently generate the dynamic graph for each time slice to reflect the changes of spatial correlations, but they follow the paradigm of independently modeling spatio-temporal dependencies, ignoring the cross-time spatial influence. In this paper, we propose a novel cross-time dynamic graph-based deep learning model, named CDGNet, for traffic forecasting. The model is able to effectively capture the cross-time spatial dependence between each time slice and its historical time slices by utilizing the cross-time dynamic graph. Meanwhile, we design a gating mechanism to sparse the cross-time dynamic graph, which conforms to the sparse spatial correlations in the real world. Besides, we propose a novel encoder-decoder architecture to incorporate the cross-time dynamic graph-based GCN for multi-step traffic forecasting. Experimental results on three real-world public traffic datasets demonstrate that CDGNet outperforms the state-of-the-art baselines. We additionally provide a qualitative study to analyze the effectiveness of our architecture. 
\end{abstract}




\maketitle

\section{Introduction}
\begin{figure}[t]
    \vspace{2mm}
    \centering
        \begin{subfigure}{0.48\linewidth}
        \includegraphics[width=\linewidth]{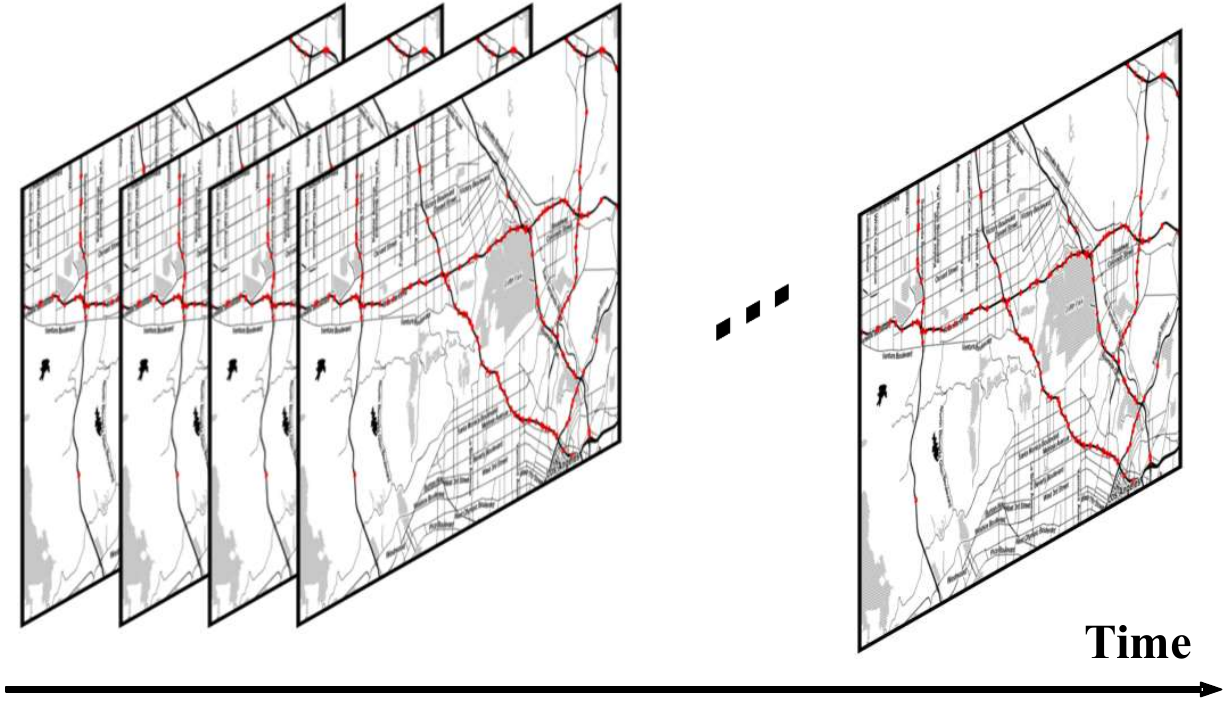}
        \captionsetup{font=small}
        \caption{Pre-defined static graph.}
        \label{psg}
      \end{subfigure}
      \hfill
      \begin{subfigure}{0.48\linewidth}
        \includegraphics[width=\linewidth]{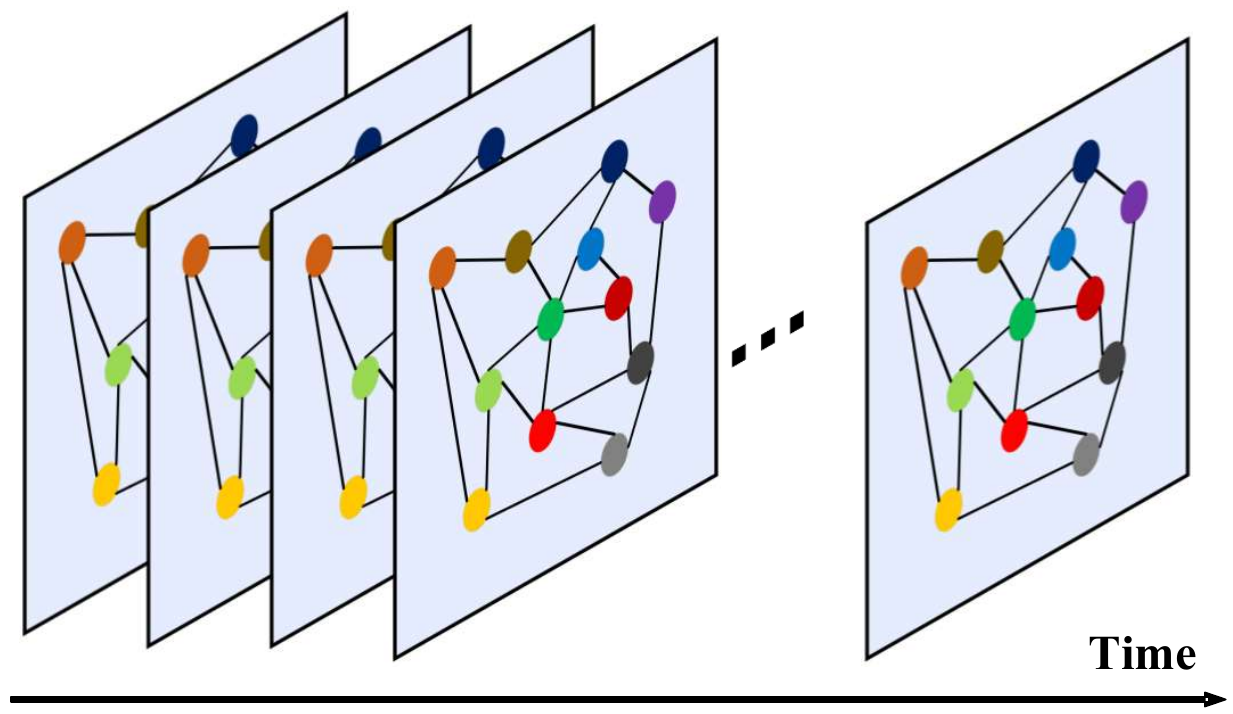}
        \captionsetup{font=small}
        \caption{Learnable static graph.}
        \label{lsg}
      \end{subfigure}
      
      \begin{subfigure}{0.48\linewidth}
        \imagebox{22mm}{\includegraphics[width=\linewidth,height=2.2cm]{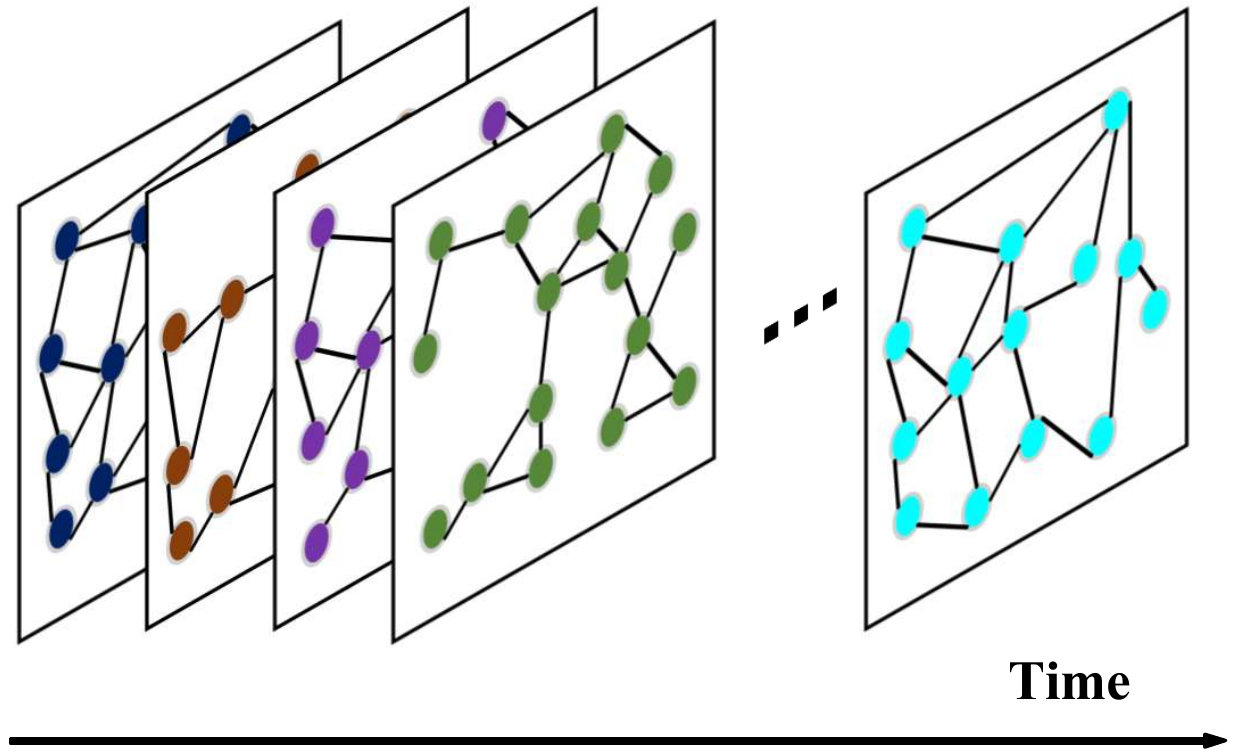}}
        \captionsetup{font=small}
        \caption{Dynamic graph.}
        \label{dg}
      \end{subfigure}
      \hfill
      \begin{subfigure}{0.48\linewidth}
        \imagebox{22mm}{\includegraphics[width=\linewidth,height=2.2cm]{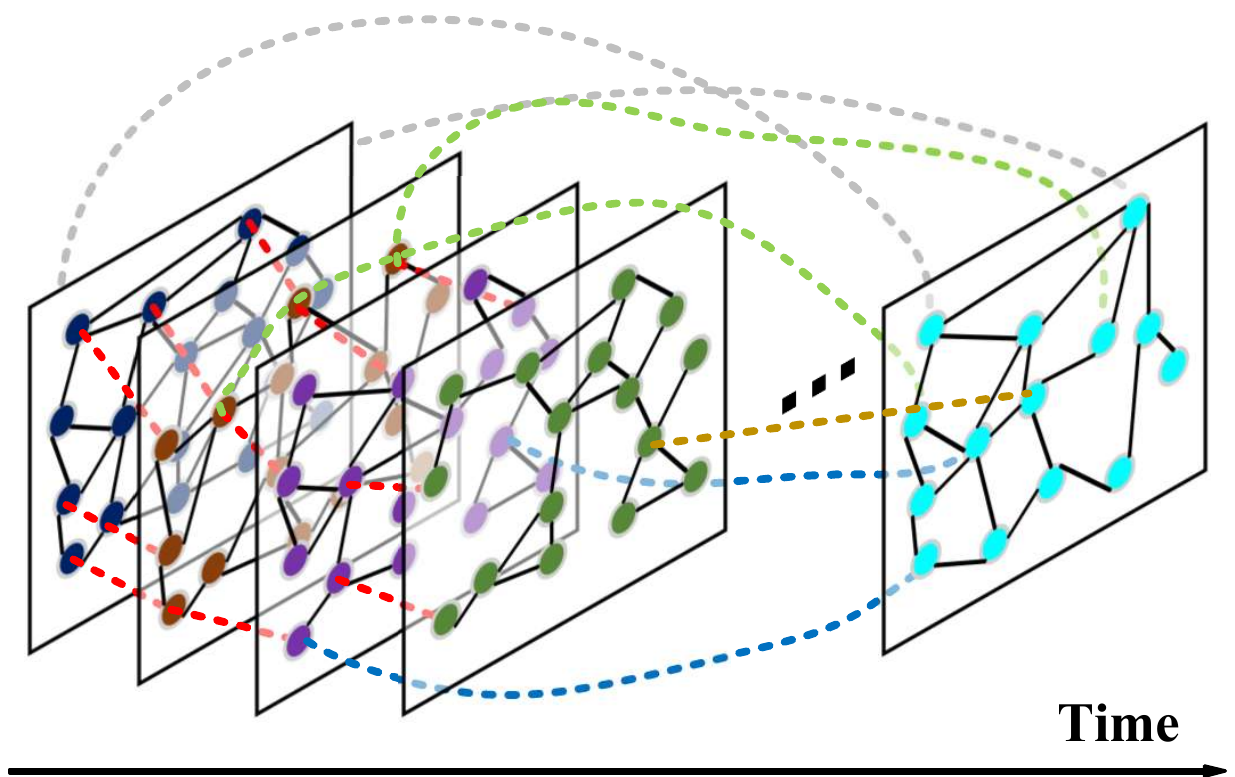}}
        \captionsetup{font=small}
        \caption{Cross-time dynamic graph.}
        \label{cdg}
      \end{subfigure}
      \caption{Graph generation methods.}
      \vspace{-12pt}
      \label{graph_gen}
\end{figure}

Traffic forecasting is important for traffic management and dispatch, which is the most challenging task in Intelligent Transportation Systems (ITS) of webs \cite{dimitrakopoulos2010intelligent}. The task aims to predict the traffic condition, \emph{e.g.}, traffic speed, volume, and consumption of vehicles, through the recorded historical traffic states \cite{DBLP:conf/iclr/LiYS018,DBLP:conf/aaai/LiZ21,elmi2021deepfec}. Traffic forecasting is challenging because of its complex and dynamic spatio-temporal dependencies in real-world road networks. Researchers dive into the task for decades and many methods are proposed. Traditional statistical methods such as Auto-Regressive Integrated Moving Average (ARIMA) \cite{williams2003modeling} and Kalman filtering \cite{wang2005real} usually rely on the stationary assumption, which is often violated by the traffic data. Shallow machine learning methods \cite{wu2004travel,van2012short} and some deep learning methods \cite{fu2016using,zhao2017lstm} capture the temporal dependence of sensors individually and ignore the complex spatial dependence. Consequently, \cite{zhang2016dnn,zhang2017deep} integrate the convolution neural networks (CNNs) \cite{gu2018recent} into recurrent neural networks (RNNs) \cite{medsker2001recurrent} to capture spatio-temporal dependencies simultaneously. However, CNNs restrict the model to process 2D-grid structures (\emph{e.g.}, images and videos), and do not consider non-Euclidean correlations dominated by irregular road networks. Graph convolution networks (GCNs) \cite{DBLP:conf/iclr/KipfW17} achieve splendid success in handling non-Euclidean data. As shown in Figure \ref{psg}, \ref{lsg}, and \ref{dg}, we divide existing graph-based models into three categories according to the graph generation style, and all of them follow the paradigm of independently modeling spatio-temporal dependencies, \emph{i.e.}, they use graph neural networks to capture the internal spatial dependence of each time slice and univariate time series networks to capture the temporal dependence of each sensor. 1) Pre-defined static graph generated according to the fixed prior graph knowledge and used for all time slices. \cite{DBLP:conf/ijcai/YuYZ18,DBLP:conf/iclr/LiYS018,DBLP:conf/aaai/LiZ21} combine the pre-defined static graph-based GCN with univariate time series models such as temporal convolution networks (TCNs) \cite{bai2018empirical} and RNNs to perform spatio-temporal prediction on non-Euclidean data. 2) Learnable static graph generated from the back-propagation through the data-driven manner is used for all time slices. \cite{DBLP:conf/nips/0001YL0020,wu2020connecting} use the learned graph to avoid the bias caused by prior knowledge. 3) Dynamic graph generated according to the temporal feature of each time slice. \cite{zheng2020gman} uses the spatial self-attention to dynamically calculate edge weights for the graph of each time slice and \cite{han2021dynamic} learns the graph for each time slice.

Although dynamic graph-based methods achieve outstanding performance in traffic forecasting, they still have some shortcomings. 1) They calculate correlations between each pair of sensors and generate the complete graph to capture the spatial dependence, which is contrary to sparse spatial correlations of road networks in the real world \cite{oreshkin2020fc}. Thus, messages will transmit by two unrelated sensors, which brings negative samples of edge to GCNs and decrease the accuracy of the forecast. 2) The impact of events will take a certain amount of time to spread to other areas in the road network, resulting in a lag in spatial dependence. For instance, traffic jams in the morning and evening rush hours will last for a long time and will affect the surrounding roads for a period in the future. Therefore, we need to extract not only the internal spatial dependence in each time slice but also the inter-spatial dependence between each time slice and its historical time slices for traffic forecasting. However, previous methods follow the paradigm of individually capturing the spatial dependence of each time slice. They ignore the continuity and hysteresis of the spatial impact and thus fail to capture the correct spatial dependence.

In this paper, we propose a cross-time dynamic graph-based deep learning model, named CDGNet, for traffic forecasting. As shown in Figure \ref{cdg}, the cross-time dynamic graph generated by our model can capture not only the intra-spatial dependence in each time slice but also the inter-spatial dependence across different time slices. When using a cross-time graph-based GCN (CDGCN), the historical spatio-temporal dependencies can be captured, unlike using the univariate time series network to capture only the historical temporal dependence. Additionally, we design a gating mechanism to sparse the cross-time dynamic graph in our model to ensure that each sensor will not be disturbed by sensors with opposite temporal patterns during the training process, which further enhances the robustness of our model. Finally, we design a novel encoder-decoder architecture to incorporate the cross-time graph-based GCN and perform multi-step traffic forecasting. The performance of long-term and short-term prediction in our model is balanced, because our architecture does not produce accumulative errors like the dynamic decoding in RNNs and does not utilize the transform attention like GMAN \cite{zheng2020gman}. The contribution of this paper can be summarized as:
\begin{itemize}
    \item We propose the cross-time dynamic graph and utilize the cross-time dynamic graph-based GCN (CDGCN) to capture the intra and inter spatial dependence. To the best of our knowledge, CDGCN is the first network of traffic forecasting that considers the cross-time spatial dependence.
    \item CDGNet, a novel cross-time dynamic graph-based deep learning model, which integrates the CDGCN into an novel encoder-decoder architecture to forecast multi-step traffic speed.
    \item Experiments on three real-world traffic datasets are conducted to evaluate the performance of our model and verify the effectiveness of each component in our model. Results show that our model outperforms the accuracy of state-of-the-art baselines.
\end{itemize}
\section{related works}
\subsection{Graph Convolution Network}
Recently, graph convolution networks (GCNs) and subsequent variants achieve state-of-the-art results in various application areas, including but not limited to social networks \cite{he2020lightgcn}, multivariate time series forecasting \cite{DBLP:conf/nips/CaoWDZZHTXBTZ20}, and natural language processing \cite{yao2019graph}. GCNs are an efficient variant of CNNs on graphs. At first, GCNs \cite{DBLP:journals/corr/BrunaZSL13} need to calculate the eigen-decomposition of the Laplacian matrix, which consumes enormous computing resources. \cite{DBLP:conf/iclr/KipfW17,defferrard2016convolutional} propose to use the k-order polynomial approximation as the spectral filter to update graph representations. Later works \cite{abu2019mixhop,chen2020simple} use the residual connection and the feature selection to increase the depth of GCNs and delay the occurrence of over-smoothing \cite{li2018deeper}. Besides, Performer \cite{DBLP:conf/iclr/ChoromanskiLDSG21} points out that self-attention can also be represented as a complete graph convolution network.
\subsection{Traffic Forecasting}
Traffic forecasting has been studied for decades. In the early years, traditional statistical methods (\emph{e.g.}, auto-regressive integrated moving average \cite{williams2003modeling} and vector auto-regression \cite{li2018brief}) are used for traffic forecasting. However, these methods rely on the stationary assumption and violate the non-linear traffic data. Shallow machine learning methods (\emph{e.g.}, support vector regression \cite{wu2004travel} and k-nearest neighbor \cite{van2012short}) can capture non-linear dependencies, but they need hand-crafted features, which require experts to do this. As the development of deep learning, \cite{fu2016using,zhao2017lstm,li2020capsules} use simple univariate time series models such as RNNs and TCNs to capture temporal dependence of each sensor individually, ignoring the spatial correlations between sensors. Later, researchers use CNNs to extract the spatial dependence in the image-based traffic forecasting task \cite{zhang2016dnn,zhang2017deep,ma2017learning}, and CNNs are limited in the road network-based traffic forecasting task. DCRNN and STGCN \cite{DBLP:conf/ijcai/YuYZ18,DBLP:conf/iclr/LiYS018} generate a static graph according to the distance between sensors in the road network and use GCNs to capture spatial dependence. Moreover, \cite{fang2019gstnet} applies GCN to image-based traffic prediction tasks and achieves certain results, and \cite{liang2021fine} further proposes a hierarchical graph convolution network to make fine-grained predictions for higher-resolution images. Subsequently, \cite{DBLP:conf/nips/0001YL0020,wu2020connecting,lin2021rest} replace the distance-based static graph in DCRNN and STGCN with the learned static graph to avoid the bias injected by prior knowledge. Besides, \cite{pan2021autostg} performs neural architecture search on the combination of 1D convolution and graph convolution to select the best architecture for traffic prediction. For the dynamic graph aspect, GMAN and STGNN \cite{zheng2020gman,wang2020traffic} use the dot-product to generate graph for each time slice through the temporal feature. DMSTGCN \cite{han2021dynamic} learns graphs through back-propagation for one day to reduce parameters. The performance of STGNN and DMSTGCN is worse than GMAN because the STGNN does not adopt the encoder-decoder architecture and the DMSTGCN does not generate graphs for each time slice. Besides, GMAN ignores that spatial correlations of sensors in the road network are sparse and not consider the inter-spatial dependence.
\section{PROBLEM STATEMENT}
We focus on traffic speed forecasting. Let $X_t=(x_t[1],...,x_t[N])\in\mathbb{R}^{N\times 1}$ denotes the recorded speed of $N$ sensors at time step $t$, where $x_t[i]\in\mathbb{R}$ denotes the value of the $i-$th sensor at time step $t$.
\begin{definition}[Graph]
A graph is formulated as $\mathcal{G}=(\mathcal{V},\mathcal{E})$, where
$\mathcal{V}$ is the set of nodes, and $\mathcal{E}$ is the set of edges. We use $N$ to denote the number of nodes in a graph.
\end{definition}
\begin{definition}[Node Neighborhood]
Let $v\in\mathcal{V}$ to denote a node and $e=(v,u)\in\mathcal{E}$ to denote an edge pointing from $v$ to $u$. The neighborhood of a node $v$ is defined as $N(v)=\{u\in\mathcal{V}\vert(v,u)\in\mathcal{E}\}$.
\end{definition}
\begin{definition}[Adjacency Matrix]
The adjacency matrix is a mathematical representation of a graph, denoted as $A\in\mathbb{R}^{N\times N}$ with $A_{v,u}=c>0$ if $(v,u)\in\mathcal{E}$ and $A_{v,u}=0$ if $(v,u)\notin\mathcal{E}$ .
\end{definition}
\newtheorem{problem}{Problem}
\begin{problem}
Given the observations of $N$ sensors at historical $P$ time slices $\mathcal{X}=(X_{1},X_2,...,X_P)\in\mathbb{R}^{P\times N\times 1}$, we aim to predict the traffic speed of the next $F$ time slices for all sensors by minimizing the $L1$ loss, denoted as $\hat{\mathcal{Y}}=(\hat{Y}_{1},\hat{Y}_{2},...,\hat{Y}_{F})\in\mathbb{R}^{F\times N\times 1}$ .
\end{problem}
\section{preliminaries}
\subsection{Graph Convolution Network}
The first-order approximate GCN \cite{DBLP:conf/iclr/KipfW17} can be written as:
\begin{equation}
        X^{(l)}=\sigma(\tilde{D}^{-\frac{1}{2}}\tilde{A}\tilde{D}^{-\frac{1}{2}}X^{(l-1)}W)\quad ,
\end{equation}
where $\sigma$ denotes the activation function and $W$ denotes the parameter of projection. $\tilde{A}=A+I$ represents the adjacency matrix with the identity matrix $I$, and $\tilde{D}$ is the degree matrix of $\tilde{A}$. Besides, the symmetric transition matrix $\tilde{D}^{-\frac{1}{2}}\tilde{A}\tilde{D}^{-\frac{1}{2}}$ can be replaced with the random walk transition matrix $\tilde{D}^{-1}\tilde{A}$ \cite{klicpera2019diffusion}. Formulated as:
\begin{equation}
        X^{(l)}=\sigma(\tilde{D}^{-1}\tilde{A}X^{(l-1)}W)\quad .
\end{equation}
\subsection{Spatial \& Temporal Self-Attention}
Self-attention \cite{vaswani2017attention} has been successful in many fields, such as natural language processing \cite{DBLP:conf/naacl/DevlinCLT19} and computer vision \cite{liu2021swin}, which uses the scaled dot-product to calculate the correlation matrix between all pairs of queries and keys, then updates values by the correlation matrix. Besides, self-attention can be rewritten as a dynamic graph convolution form \cite{DBLP:conf/iclr/ChoromanskiLDSG21}:
\begin{equation}
    \begin{split}
        &Attention(Q,K,V)=D^{-1}AV\\
        &where\quad A=exp(\frac{QK^T}{\sqrt{d}})\quad ,\\
    \end{split}
\end{equation}
where $exp(\cdot)$ is the exponential operation and $d$ is the feature dimension of inputs. $D$ is the degree matrix of $A$.

Multi-head self-attention is the most widely adopted self-attention in practice, which means to jointly attend to information from different representation subspaces, formally:
\begin{equation}
    \begin{split}
        &MultiHead(Q,K,V)=Concat(head_1,...,head_h)W^O\\
        &where\quad head_i=Attention(QW_i^Q,KW_i^K,VW_i^V)\quad ,\
    \end{split}
\end{equation}
where $Concat(\cdot)$ indicates the concatenate operation. $W_i^Q\in\mathbb{R}^{d\times \frac{d}{h}}$, $W_i^K\in\mathbb{R}^{d\times \frac{d}{h}}$, $W_i^V\in\mathbb{R}^{d\times \frac{d}{h}}$, and $W^O\in\mathbb{R}^{d\times d}$ are parameter matrices of projections.

GMAN \cite{zheng2020gman} applies the self-attention to the temporal and spatial dimension to dynamically capture the temporal dependence for each sensor and the spatial dependence for each time slice. The temporal and spatial self-attention can be formulated as:
\begin{equation}
    \begin{split}
        &TAtt=Concat(ta_1,...,ta_n,...,sa_N)\\
        &where\quad ta_n=MultiHead(X[n],X[n],X[n])\quad ,
    \end{split}
\end{equation}
where $X[n]\in\mathbb{R}^{T\times d}$ indicates the traffic data of sensor $n$. Generally, a causal mask matrix is additionally used in $TAtt(\cdot)$ to avoid receiving future information. When the input is the historical data, $T=P$; and when the input is the predicted data, $T=F$.
\begin{equation}
    \begin{split}
        &SAtt=Concat(sa_1,...,sa_t,...,sa_T)\\
        &where\quad sa_t=MultiHead(X_t,X_t,X_t)\quad ,
    \end{split}
\end{equation}
where $X_t\in\mathbb{R}^{N\times d}$ indicates the traffic data at time $t$.
\begin{figure*}[t]
  \centering
  \includegraphics[width=\linewidth]{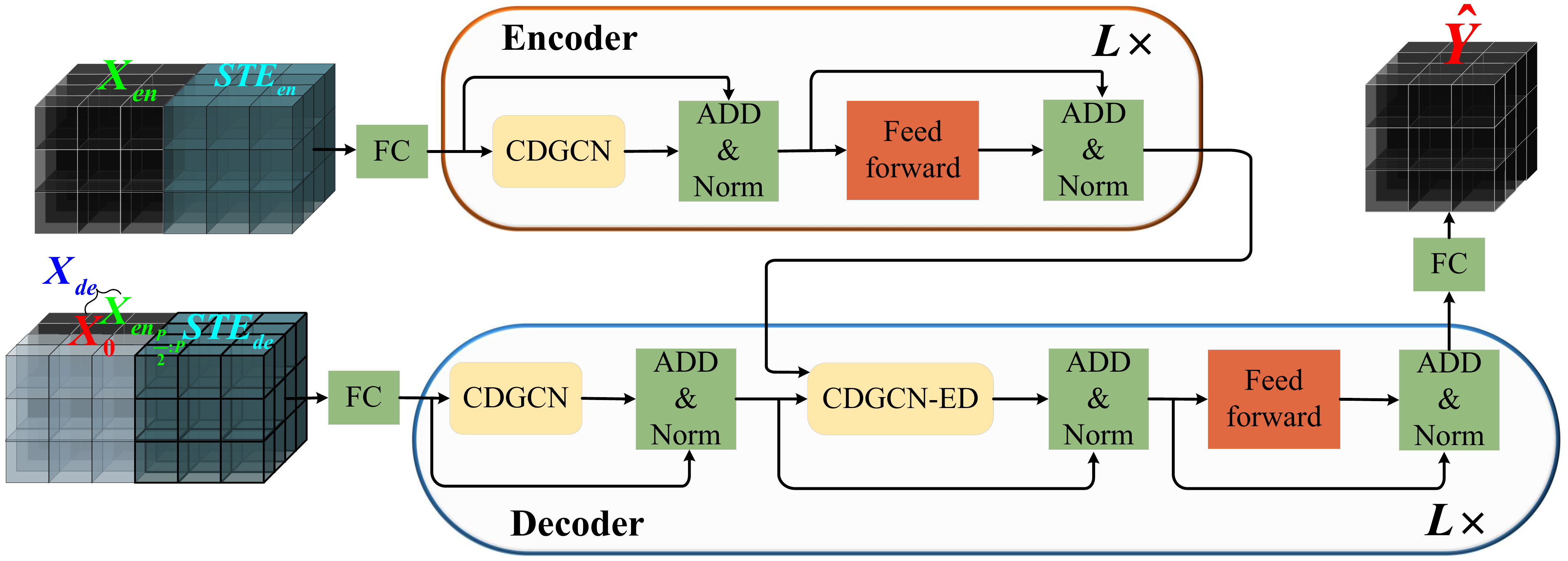}
  \caption{The overview of CDGNet, which consists of an encoder and a decoder both with $L$ layers, and three fully-connected layers (FC). $\mathcal{X}_{en}\in\mathbb{R}^{P\times N\times 1}$ and $\mathcal{X}_{de}\in\mathbb{R}^{(\frac{P}{2}+F)\times N\times 1}$ are the inputs of encoder and decoder respectively, where $\mathcal{X}_{de}$ comprises of $\mathcal{X}_{en_{\frac{P}{2}:P}}$ and $\mathcal{X}_0\in\mathbb{R}^{F\times N\times 1}$. Moreover, $STE_{en}\in\mathbb{R}^{P\times N\times d}$ and $STE_{de}\in\mathbb{R}^{(\frac{P}{2}+F)\times N\times d}$ are the corresponding spatio-temporal embedding. Besides, ADD and Norm indicates the residual connection and layer normalization.}
  \label{model}
\end{figure*}
\section{Methodology}
Figure \ref{model} illustrates the framework of our CDGNet, which adopts an encoder-decoder architecture for traffic forecasting. Each layer in encoder is composed of the cross-time dynamic graph-based graph convolution network (CDGCN) and the feed forward sub-layer to capture spatio-temporal dependencies from history. Compared with encoder, in addition to using the CDGCN and feed forward sub-layer to capture spatio-temporal dependencies from predicted sequences, decoder further uses the encoder-decoder cross-time dynamic graph-based graph convolution network (CDGCN-ED) to extract historical information from the output of encoder.
\subsection{Input Layer}
As shown in Figure \ref{model}, we use fully connected layers before traffic data $\mathcal{X}\in\mathbb{R}^{P\times N\times 1}$ enters the model to transform the data from traffic speed to high-dimension space to make our model more robust, formally:
\begin{equation}
    \mathcal{X}_{en}^0=ReLU(W^{I_1}\mathcal{X}+b^{I_1})W^{I_2}+b^{I_2}\quad ,
\end{equation}
where $W^{I_1}\in\mathbb{R}^{1\times d}$, $b^{I_1}\in\mathbb{R}^{d}$, $W^{I_2}\in\mathbb{R}^{d\times d}$, and $b^{I_2}\in\mathbb{R}^{d}$ are learnable parameters. $d$ is the dimension of the feature space.
\subsection{Encoder and Decoder Stacks}
\subsubsection{Encoder}
The encoder consists of $L$ stacked identical layers, each layer includes two sub-layers. The first is our CDGCN, and the second is a simple fully connected feed-forward network. We adopt residual connection for each sub-layer and employ layer normalization to normalize each sub-layer. The overall equations for $l$-th encoder layer are summarized as $\mathcal{X}_{en}^l= Encoder(Concat(\mathcal{X}_{en}^{l-1},STE_{en}))$, where $STE_{en}\in\mathbb{R}^{P\times N\times d}$ is the spatio-temporal embedding of the graph structure and time information. Details are shown as follows:
\begin{equation}
\begin{split}
    \mathcal{D}_{en}^{l}&=LayerNorm(CDGCN(Concat(\mathcal{X}_{en}^{l-1},STE_{en}))+\mathcal{X}_{en}^{l-1})\\
    \mathcal{X}_{en}^l&=LayerNorm(FeedForward(\mathcal{D}_{en}^l)+\mathcal{D}_{en}^l)\quad ,\\
\end{split}
\end{equation}
where $CDGCN(\cdot)$ denotes the cross-time dynamic graph-based graph convolution and we will give a detailed description of it in the next section.
\subsubsection{Decoder}
Each layer in decoder contains the CDGCN and CDGCN-ED, which can refine the prediction and utilize the past spatio-temporal information, respectively. Suppose there are $L$ layers in decoder. With the hidden states $\mathcal{X}_{en}^{L}$ from encoder, the equation of $l$-th layer in decoder can be summarized as $\mathcal{X}_{de}^l= Decoder(Concat(\mathcal{X}_{de}^{l-1},STE_{de}),\mathcal{X}_{en}^L)$, details are:
\begin{equation}
\begin{split}
    \mathcal{D}_{de}^{l_1}&=LayerNorm(CDGCN(Concat(\mathcal{X}_{de}^{l-1},STE_{de}))+\mathcal{X}_{de}^{l-1})\\
    \mathcal{D}_{de}^{l_2}&=LayerNorm(CDGCN-ED(\mathcal{D}_{de}^{l_1},\mathcal{X}_{en}^L)+\mathcal{D}_{de}^{l_1})\\
    \mathcal{X}_{de}^l&=LayerNorm(FeedForward(\mathcal{D}_{de}^{l_2})+\mathcal{D}_{de}^{l_2})\quad .\\
\end{split}
\end{equation}
where $\mathcal{X}_{de}^0=Concat(\mathcal{X}_{en_{\frac{P}{2}:P}}^0,\mathcal{X}_0)\in\mathbb{R}^{(\frac{P}{2}+F)\times N\times d}$. $\mathcal{X}_0\in\mathbb{R}^{F\times N\times d}$ denotes placeholders filled with zero. Dimension $d$ of all layers in our model is $64$ in order to facilitate the residual connection.
\subsection{Output layer}
Similarly with encoder, we use fully connected layers to convert the output of decoder from high-dimension space to the traffic speed:
\begin{equation}
    \hat{\mathcal{Y}}=ReLU(\mathcal{X}_{de}^LW^{O_1}+b^{O_1})W^{O_2}+b^{O_2}\quad ,
\end{equation}
where $W^{O_1}\in\mathbb{R}^{d\times d}$, $W^{O_2}\in\mathbb{R}^{d\times 1}$, $b^{O_1}\in\mathbb{R}^d$, and $b^{O_2}\in\mathbb{R}$ are learnable parameters.
\subsection{Cross-time Dynamic Graph-based Graph Convolution Network}
It is important to capture the dynamic spatial dependence in the road network, but existing dynamic graph-based traffic forecasting methods, such as \cite{zheng2020gman}, calculate the dense adjacency matrix at each time slice by the dot-product operation, ignoring the sparsity of spatial correlations. Therefore, we design a gating mechanism to sparse the dense adjacency matrix obtained from the dot-product and add a identity matrix to the sparse adjacency matrix to enhance the self-expression ability, formally:
\begin{equation}
    \tilde{A}=ReLU(QK^T)+I\quad ,
\end{equation}
where $ReLU(\cdot)$ is the non-linear activation function.

Besides, the exponential operation in self-attention can cause the gradient to disappear or explode, so we normalize the sparse adjacency matrix by dividing its degree matrix $\tilde{D}\in\mathbb{R}^{N\times N}$. The dynamic graph convolution network in our model is:
\begin{equation}
    DGCN(Q,K,V)=\tilde{D}^{-1}\tilde{A}V\quad .
\end{equation}

We further extend the DGCN to multi-head to attend information from different subspaces:
\begin{equation}
    \begin{split}
        &MHDGCN(Q,K,V)=Concat(head_1,...,head_h)W^O\\
        &where\quad head_i=DGCN(QW_i^Q,KW_i^K,VW_i^V)\quad ,\\
    \end{split}
\end{equation}
where $W_i^Q\in\mathbb{R}^{d\times \frac{d}{h}}$, $W_i^K\in\mathbb{R}^{d\times \frac{d}{h}}$, $W_i^V\in\mathbb{R}^{d\times \frac{d}{h}}$ are the projection parameters, and $W^O\in\mathbb{R}^{d\times d}$ is the parameter of the final projection.

\begin{figure}
    \centering
    \includegraphics[width=1.0\linewidth]{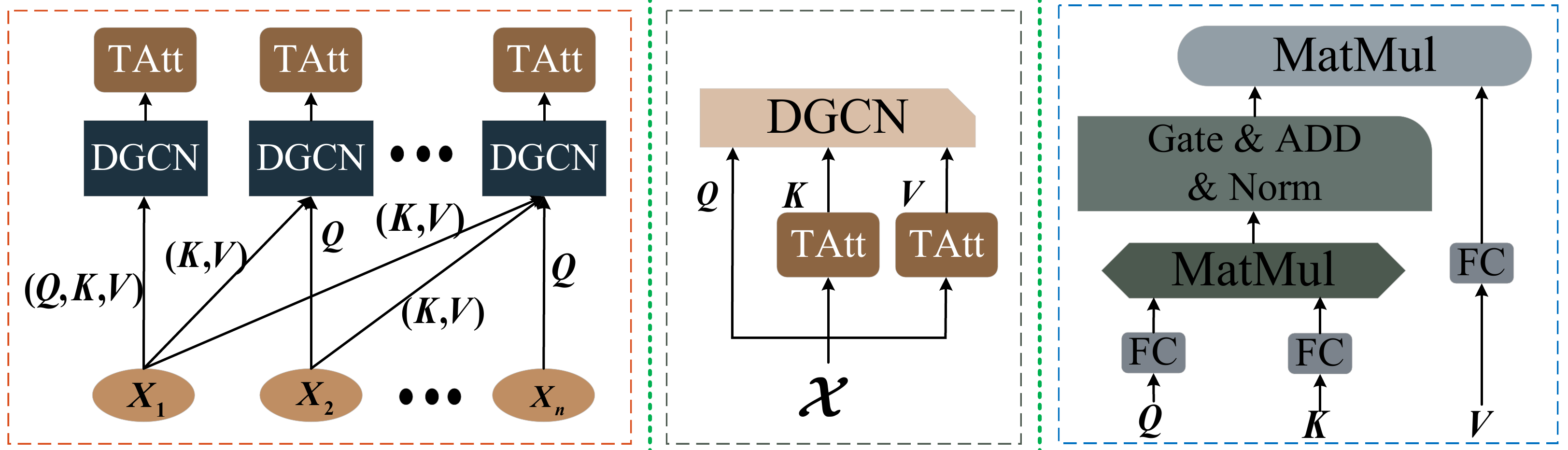}
    \caption{The left is the original CDGCN, the center is the final CDGCN, and the right is the used DGCN in CDGCN. Gate, ADD, and Norm indicates the gating mechanism, add identity matrix, and degree normalization.}
    \vspace{-13pt}
    \label{graph_methods}
\end{figure}

On the other hand, prior dynamic graph-based methods ignore that two dimensions of temporal and spatial in the road network are not isolated but related. For instance, when accidents occur in the road network, it does not affect all nodes in time but has a hysteresis. In other words, the farther away the node is from the accident, the slower it will be affected, just like the transmission of sound waves takes time. Using spatial self-attention in isolation for each time slice without considering the spatial impact along with the temporal dimension cannot capture the inter-spatial dependence caused by message passing in the traffic system. Therefore, we propose a novel cross-time dynamic graph based graph convolution network to capture the inter-spatial dependence. As shown in the left of Figure \ref{graph_methods}, DGCN is used not only for each time slice but also between each time slice and their previous time slices to capture inter-spatial dependence in the original CDGCN, formally:
\begin{equation}
    \begin{split}
        &CDGCN_{Original}=Concat(cdgcn_1,...,cdgcn_t,...,cdgcn_T)\\
        &where\quad cdgcn_t=TAtt(MHDGCN(X_t,\mathcal{X}_{1:t},\mathcal{X}_{1:t}))\quad ,\\
    \end{split}
\end{equation}
where $X_t\in\mathbb{R}^{N\times d}$ and $\mathcal{X}_{1:t}\in\mathbb{R}^{t\times N\times d}$ represent the data at time $t$ and all the data before time $t$, respectively. We use $TAtt(\cdot)$ in Equation 14 to the weighted sum for the states with historical spatial dependencies for each time slice.
\subsubsection{Efficient Computation}
Although we can use the original CDGCN to capture the inter-spatial dependence across time, its complexity is $O(T^5N^3d^2)$ and is difficult to parallelize because of the nested two layers of loops. The consumption of the original CDGCN is unaffordable for us. As shown in the center of Figure \ref{graph_methods}, we reverse the original process to simplify CDGCN, \emph{i.e.}, we move $TAtt(\cdot)$ in Equantion 14 into $MHDGCN(\cdot)$. Specifically, we first use $TAtt(\cdot)$ to sum the key and value with historical information, and then perform dynamic graph convolution on the compressed key and value to obtain the inter-spatial dependence. The complexity of the final version of our CDGCN is reduced to $O(T^3N^3d^2)$. In addition, the final CDGCN can be processed in parallel to reduce time consumption. The final CDGCN can be formalized as follows:
\begin{equation}
    CDGCN=MHDGCN(\mathcal{X},TAtt(\mathcal{X}),TAtt(\mathcal{X}))\quad .
\end{equation}
\subsubsection{Encoder-Decoder CDGCN}
In order to interact with historical information, we utilize the encoder-decoder CDGCN (\emph{i.e.}, CDGCN-ED) in the decoder. The difference between CDGCN-ED and CDGCN is that the key and value of them are the output of the encoder and the output of the last sub-layer. Moreover, because all historical information should be received in the predicted sequence, $TAtt(\cdot)$ in CDGCN-ED no longer uses the causal mask matrix.
\subsection{Other Components}
\subsubsection{Feed-Forward Networks}
Each layer in our encoder-decoder architecture contains a fully connected feed-forward network, which consists of two linear transformations with a $ReLU(\cdot)$ function.
\begin{equation}
    FeedForward=ReLU(\mathcal{X}W^{F_1}+b^{F_1})W^{F_2}+b^{F_2}\quad ,
\end{equation}
where $W^{F_1}$, $W^{F_2}\in\mathbb{R}^{d\times d}$, $b^{F_1}$, $b^{F_2}\in\mathbb{R}^d$ are learnable parameters.
\subsubsection{Spatial-Temporal Embedding}
In order to more effectively distinguish sensors in different times, we concatenate the spatio-temporal embedding ($STE$) with input before each layer so that the model can take graph structure and time information into account. The spatial embedding matrix is a learnable matrix $SE\in\mathbb{R}^{N\times d_{se}}$, which is initialed with the matrix generated by node2vec \cite{grover2016node2vec}. The temporal embedding matrix $TE\in\mathbb{R}^{(P+F)\times d_{te}}$ generated by one-hot encoding. Then we fed the spatial and temporal embedding metrics into two fully connected layers and derive the $SE\in\mathbb{R}^{N\times d}$ and $TE\in\mathbb{R}^{(P+F)\times d}$, respectively. To obtain the time-variant vertex representations, we fuse the aforementioned spatial embedding and temporal embedding as $STE\in\mathbb{R}^{(P+F)\times N\times d}$.
\subsection{Loss Function}
We can train our model end-to-end via back-propagation by minimizing the $L1$ loss between predicted values and ground truths:
\begin{equation}
    \mathcal{L}(\Theta)=\frac{1}{F\times N}\sum_{t=1}^F\sum_{n=1}^N\vert \hat{Y}_{t}[n]-X_{t}[n]\vert\quad ,
\end{equation}
where $\Theta$ indicates all the parameters in our model.
\begin{table}[t]
    \centering
    \caption{Dataset statistics.}
    \resizebox{.47\textwidth}{!}{\begin{tabular}{lcccc}
    \toprule
    Datasets & \#Nodes & \#Edges & \#TimeSlices & Timespan\\
    \midrule
    METR-LA & 207 & 1515 & 34272 & 3/1/2012-6/30/2012\\
    PEMS-BAY & 325 & 2369  & 52116  & 1/1/2017-5/31/2017\\
    PEMSD4 & 307 & 340 & 16992 & 1/1/2018-2/28/2018\\
    \bottomrule
    \end{tabular}}
    \label{sta}
\end{table}
\subsection*{Model Complexity Analysis}
The complexity of the vanilla spatial and temporal self-attention is $O(TN^2d)$ and $O(NT^2d)$, respectively. For the original CDGCN, we utilize the DGCN between each time slice and their history time slices; the process requires two layers of loops and cannot be processed in parallel, and we need a temporal self-attention to aggregate historical spatial information. The complexity of original CDGCN is $O(T^5N^3d^2)$. We further remove the double loops in the original CDGCN and simply stack temporal and spatial self-attention to capture the inter-spatial dependence across time. Therefore, the complexity of final CDGCN is $O(T^3N^3d^2)$.
\section{evaluation}
\subsection{Experimental Settings}
\subsubsection{Datasets}
We conduct experiments on three real-world traffic datasets, which are METR-LA, PEMS-BAY, and PEMSD4 \cite{DBLP:conf/iclr/LiYS018,DBLP:conf/aaai/LiZ21}. Statistics of these datasets are summarized in Table \ref{sta}. METR-LA and PEMS-BAY are divided into a training set (70\%), validation set (10\%), and test set (20\%) in chronological order. PEMSD4 is divided into a training set (60\%), validation set (20\%), and test set (20\%) in chronological order. For both datasets, we utilize the history of 12 horizons to predict the next 12 horizons (1 hour) and use Z-score normalization to preprocess input data.
\subsubsection{Metrics} We use three evaluation metrics, including Mean Absolute Error (MAE), Rooted Mean Square Error (RMSE), and Mean Absolute Percentage Error (MAPE). Note that the lower value of these metrics represents the higher prediction accuracy.
\begin{table*}[t]
\aboverulesep=0ex
\belowrulesep=0ex
\centering
\caption{Traffic forecasting performance comparison of CDGCN and other baseline models. }
\resizebox{1.0\linewidth}{!}{\begin{tabular}{l|l|ccc|ccc|ccc|ccc} 
\toprule[1.5pt]
\multirow{2}{*}{Datasets} & \multirow{2}{*}{Methods} & \multicolumn{3}{c|}{Horizon 3} & \multicolumn{3}{c|}{Horizon 6} & \multicolumn{3}{c|}{Horizon 12} & \multicolumn{3}{c}{Average}\\
\cmidrule{3-14}
    &  & MAE & RMSE & MAPE ($\%$) & MAE & RMSE & MAPE ($\%$) & MAE & RMSE & MAPE ($\%$) & MAE & RMSE & MAPE ($\%$)\\
\midrule[1pt]
\multirow{10}{*}{METR-LA}
& DCRNN & 2.79 & 5.82 & 7.20 & 3.16 & 6.44 & 8.63 & 3.61 & 7.50 & 10.39 & 3.15 & 6.56 & 8.67\\
& STGCN & 2.87 & 5.89 & 7.57 & 3.49 & 7.37 & 9.56 & 4.52 & 9.48 & 12.37 & 3.69 & 7.43 & 9.67\\
& STFGNN & 2.77 & 5.62 & 7.33 & 3.15 & 6.35 & 8.75 & 3.65 & 7.47 & 10.49 & 3.18 & 6.40 & 8.81\\
& AGCRN & 2.78 & 5.74 & 7.48 & 3.10 & 6.43 & 8.52 & 3.50 & 7.49 & 9.99 & 3.11 & 6.47 & 8.54\\
& MTGNN & 2.69 & 5.30 & 7.12 & 3.07 & 6.33 & 8.39 & 3.51 & 7.42 & 9.92 & 3.06 & 6.35 & 8.38\\
\cmidrule{2-14}
& STGNN & 2.74 & 5.57 & 7.63 & 3.15 & 6.60 & 9.28 & 3.58 & 7.62 & 10.96 & 3.10 & 6.52 & 9.08\\
& DMSTGCN-P & 2.69 & 5.29 & 7.11 & 3.06 & 6.33 & 8.38 & 3.49 & 7.40 & 9.91 & 3.05 & 6.50 & 8.36\\
& GMAN & 2.80 & 5.53 & 7.37 & 3.10 & 6.42 & 8.39 & 3.44 & 7.32 & 9.99 & 3.08 & 6.41 & 8.31\\
\cmidrule{2-14}
& CDGNet & \textbf{2.66} & \textbf{5.19} & \textbf{6.91} & \textbf{2.98} & \textbf{6.10} & \textbf{8.22} & \textbf{3.32} & \textbf{7.03} & \textbf{9.71} & \textbf{2.94} & \textbf{6.03} & \textbf{8.12}\\
\midrule[1pt]
\multirow{10}{*}{PEMS-BAY}
& DCRNN & 1.39 & 2.96  & 2.92 & 1.76  & 3.99 & 3.93 & 2.09 & 4.76 & 4.95 & 1.75 & 3.92 & 3.93\\
& STGCN & 1.36 & 2.97 & 2.89 & 1.83 & 4.26 & 4.18 & 2.46 & 5.66 & 5.81 & 1.87 & 4.28 & 4.30\\
& STFGNN & 1.36 & 2.81 & 2.83 & 1.67 & 3.79 & 3.78 & 1.97 & 4.52 & 4.64 & 1.66 & 3.74 & 3.77\\
& AGCRN & 1.40 & 3.02 & 3.13 & 1.66 & 3.89 & 3.83 & 1.96 & 4.54 & 4.49 & 1.67 & 3.85 & 3.72\\
& MTGNN & 1.33 & 2.80 & 2.79 & 1.64 & 3.77 & 3.70 & 1.94 & 4.50 & 4.55 & 1.64 & 3.73 & 3.70\\
\cmidrule{2-14}
& STGNN & 1.38 & 2.99 & 2.84 & 1.69 & 3.91 & 3.77 & 1.96 & 4.59 & 4.60 & 1.70 & 3.78 & 3.85\\
& DMSTGCN-P & 1.34 & 2.82 & 2.81 & 1.65 & 3.75 & 3.72 & 1.94 & 4.47 & 4.53 & 1.64 & 3.74 & 3.69\\
& GMAN & 1.35 & 2.92  & 2.89 & 1.64  & 3.75 & 3.70 & 1.92 & 4.41 & 4.44 & 1.63 & 3.71 & 3.69\\
\cmidrule{2-14}
& CDGNet  & \textbf{1.29} & \textbf{2.77} & \textbf{2.71} & \textbf{1.59}  & \textbf{3.66} & \textbf{3.57} & \textbf{1.84} & \textbf{4.31} & \textbf{4.34} & \textbf{1.52} & \textbf{3.55} & \textbf{3.42}\\
\midrule[1pt]
\multirow{10}{*}{PEMSD4}
& DCRNN & 1.42 & 2.98 & 2.82 & 1.79 & 4.04 & 3.79 & 2.23 & 5.14 & 4.99 & 1.75 & 4.03 & 3.73\\
& STGCN & 1.42 & 2.93 & 2.78 & 1.84 & 4.02 & 3.83 & 2.39 & 5.27 & 5.13 & 1.81 & 4.05 & 3.76\\
& STFGNN & 1.40 & 2.93 & 2.77 & 1.72 & 3.87 & 3.66 & 2.07 & 4.70 & 4.60 & 1.68 & 3.80 & 3.56\\
& AGCRN & 1.43 & 3.13 & 3.01 & 1.69 & 3.87 & 3.64 & 2.02 & 4.73 & 4.47 & 1.68 & 3.87 & 3.60\\
& MTGNN & 1.35 & 2.88 & 2.69 & 1.67 & 3.82 & 3.59 & 2.01 & 4.68 & 4.57 & 1.66 & 3.79 & 3.58\\
\cmidrule{2-14}
& STGNN & 1.41 & 3.01 & 2.74 & 1.72 & 3.96 & 3.64 & 1.96 & 4.63 & 4.44 & 1.69 & 3.89 & 3.65\\
& DMSTGCN-P & 1.34 & 2.89 & 2.68 & 1.65 & 3.80 & 3.52 & 1.95 & 4.59 & 4.33 & 1.64 & 3.75 & 3.45\\
& GMAN & 1.42 & 3.09 & 2.86 & 1.65 & 3.81 & 3.50 & 1.91 & 4.56 & 4.28 & 1.63 & 3.77 & 3.44\\
\cmidrule{2-14}
& CDGNet & \textbf{1.30} & \textbf{2.82} & \textbf{2.57} & \textbf{1.59}  & \textbf{3.70} & \textbf{3.33} & \textbf{1.85} & \textbf{4.40} & \textbf{4.08} & \textbf{1.54} & \textbf{3.61} & \textbf{3.22}\\
\bottomrule[1.5pt]
\end{tabular}}
\label{main_res}
\end{table*}
\begin{table}[t]
\aboverulesep=0ex
\belowrulesep=0ex
    \centering
    \caption{Transfer performance of architecture on the METR-LA dataset with MAE metric.}
    \vspace{-8pt}
    \resizebox{1.0\linewidth}{!}{\begin{tabular}{l|l|llll}
    \toprule[1.5pt]
        Datasets & Methods & Horizon 3 & Horizon 6 & Horizon 12 & Average \\
    \midrule[0.8pt]
        \multirow{2}{*}{METR-LA} & GMAN (Original) & 2.80 & 3.10 & 3.44 & 3.08\\
        & GMAN (ours) & 2.74 & 3.07 & 3.41 & 3.02\\
        \midrule[0.8pt]
        \multirow{2}{*}{PEMS-BAY} & GMAN (Original) & 1.35 & 1.64 & 1.92 & 1.63\\
        & GMAN (ours) & 1.32 & 1.63 & 1.90 & 1.57\\
        \midrule[0.8pt]
        \multirow{2}{*}{PEMSD4} & GMAN (Original) & 1.42 & 1.65 & 1.91 & 1.63\\
        & GMAN (ours) & 1.33 & 1.63 & 1.91 & 1.58\\
    \bottomrule[1.5pt]
    \end{tabular}}
    \label{arc}
    \vspace{-12pt}
\end{table}
\subsubsection{Baselines} We compare our model with following baselines:
\begin{itemize}[leftmargin=*]
    \item \textbf{DCRNN} \cite{DBLP:conf/iclr/LiYS018}: Diffusion convolution recurrent neural network, which integrates the static distance graph-based diffusion convolution in recurrent neural networks to capture spatio-temporal dependencies simultaneously.
    \item \textbf{STGCN} \cite{DBLP:conf/ijcai/YuYZ18}: Spatial-temporal graph convolution network, which incorporates the static distance graph-based graph convolution with 1D convolutions to capture spatio-temporal dependencies.
    \item \textbf{STFGNN} \cite{DBLP:conf/aaai/LiZ21}: Spatial-temporal fusion graph neural networks, which propose a novel temporal graph to capture more spatial correlations calculated by the dynamic time warping algorithm.
    \item \textbf{AGCRN} \cite{DBLP:conf/nips/0001YL0020}: Adaptive graph convolution recurrent network is based on DCRNN. It learns a more complete posterior graph through back-propagation. In addition, it proposes a method to accelerate graph convolution.
    \item \textbf{MTGNN} \cite{wu2020connecting}: A general graph neural network framework, which combines learnt graph based mix-hop with dilated inception to extract spatio-temporal dependencies without no prior knowledge. Besides, it proposes a novel curriculum learning method for multi-step traffic forecasting.
    \item \textbf{STGNN} \cite{wang2020traffic}: A model that combines GCN, GRU, and temporal self-attention. At the same time, it uses the spatial self-attention to dynamically calculate graphs for each time slice.
    \item \textbf{DMSTGCN-P} \cite{han2021dynamic}: It is based on STGCN and learns the posterior graph for one day through back-propagation. Besides, it uses traffic volume to assist speed prediction. To be fair, we only use the primary feature of traffic speed in the comparison.
    \item \textbf{GMAN} \cite{zheng2020gman}: A graph multi-attention network, which uses spatial self-attention and temporal self-attention to capture dynamic spatio-temporal dependencies, respectively, and utilizes a gated fusion to fuse results after spatio-temporal self-attention.
\end{itemize}
\subsubsection{Hyper-Parameters and Other Settings} All of our experiments are conducted on a CentOS server (CPU: Intel(R) Xeon(R) Gold 6132 CPU @ 2.60GHz, GPU: Tesla-V100). We use the Adam optimizer to train our model with $10$ epochs. The learning rate is halved every epoch after the fifth epoch, starting with $1e-3$. The batch size of our model is $16$. There are four important hyper-parameters in our model, the dimension of each head is $d_h=8$, the number of heads is $h=8$, so the dimension of our model is $d=64$, and the number of layers in the encoder-decoder architecture is $L=3$. The dimension of the node embeddings from node2vec is $d_{se}=64$ and the dimension of one-hot encoding is $d_{te}=295$, following \cite{zheng2020gman}.
\begin{table*}[t]
\aboverulesep=0ex
\belowrulesep=0ex
\centering
\caption{Experimental results with weekdays and weekends on the METR-LA dataset.}
\resizebox{1.0\linewidth}{!}{\begin{tabular}{c|ccc|ccc|ccc|ccc} 
\toprule[1.5pt]
\multirow{2}{*}{Model / T} & \multicolumn{3}{c|}{Horizon 3} & \multicolumn{3}{c|}{Horizon 6} & \multicolumn{3}{c|}{Horizon 12} & \multicolumn{3}{c}{Average}\\
\cline{2-13}
& MAE & RMSE & MAPE (\%) & MAE & RMSE & MAPE (\%) & MAE & RMSE & MAPE (\%) & MAE & RMSE & MAPE (\%)\\
\midrule[1pt]
DCRNN (Weekdays) & 2.60 & 5.17 & 6.76 & 3.01 & 6.32 & 8.13 & 3.49 & 7.51 & 9.88 & 2.97 & 6.16 & 8.00\\
MTGNN (Weekdays) & 2.63 & 5.25 & 6.78 & 3.01 & 6.27 & 8.13 & 3.44 & 7.34 & 9.70 & 2.97 & 6.13 & 7.97\\
GMAN (Weekdays) & 2.74 & 5.53 & 7.30 & 3.04 & 6.45 & 8.57 & 3.35 & 7.24 & 9.77 & 3.00 & 6.28 & 8.39\\
CDGNet (Weekdays) & \textbf{2.58} & \textbf{5.17} & \textbf{6.75} & \textbf{2.87} & \textbf{6.05} & \textbf{8.00} & \textbf{3.19} & \textbf{6.93} & \textbf{9.32} & \textbf{2.83} & \textbf{5.98} & \textbf{7.85}\\
\midrule[1pt]
DCRNN (Weekends) & 3.26 & 7.01 & 8.30 & 3.96 & 8.72 & 10.56 & 5.03 & 10.93 & 13.94 & 3.97 & 8.62 & 10.58\\
MTGNN (Weekends) & 3.03 & 5.75 & 7.94 & 3.38 & 6.77 & 9.69 & 3.76 & 7.70 & 11.24 & 3.33 & 6.71 & 9.87\\ 
GMAN (Weekends) & 3.15 & 6.13 & 8.74 & 3.51 & 7.13 & 10.20 & 3.85 & 7.97 & 11.57 & 3.46 & 6.93 & 9.99\\
CDGNet (Weekends) & \textbf{2.94} & \textbf{5.67} & \textbf{7.68} & \textbf{3.32} & \textbf{6.69} & \textbf{9.18} & \textbf{3.71} & \textbf{7.69} & \textbf{10.86} & \textbf{3.26} & \textbf{6.59} & \textbf{9.03}\\
\bottomrule[1.5pt]
\end{tabular}}
\label{week_res}
\end{table*}
\vspace{-5pt}
\subsection{Experimental Results and Analysis}
\subsubsection{Main Results}
Table \ref{main_res} provides the experimental results of CDGNet and baselines on three traffic speed datasets. We repeat each experiment 5 times and report the average of MAE, RMSE, and MAPE. We compare graph-based deep learning methods and divide them into two types according to the graph used in GCN belong to static or dynamic. DCRNN and STGCN are the first to use GCN to predict traffic speed, they generate the static graph based on the distance between sensors in the road network. STFGNN not only uses a GCN based on the static distance-based graph, but also uses a GCN based on the temporal graph created by the similarity of time series. Because the prior knowledge is incomplete or misleading, the performance of static prior knowledge-based GCN models is adequate. Although AGCRN and MTGNN replace the prior knowledge-based graph with the posterior graph learned through a data-driven manner on the basis of DCRNN and STGCN to improve the forecasting performance, they still use one adjacency matrix at all times, ignoring dynamic spatial correlations. The performance of DMSTGCN-P and GMAN on the three datasets proves the importance and correctness of using dynamic graph. However, DMSTGCN-P allows all dates to share the learned adjacency matrix of one day, but the temporal pattern of working days and holidays is very different, which results in over-fitting. Using self-attention to dynamically calculate the adjacency matrix of each time slice seems to be very good, but existing methods do not use self-attention well. First, self-attention-based models need to follow Transformer \cite{vaswani2017attention} construction, \emph{i.e.}, they must include residual connections, layer normalization, and feed-forward. STGNN uses spatial self-attention to generate the dynamic graph, but performance is poor because it does not adopts a Transformer architecture. Second, using transform attention between in encoder and decoder like GMAN is not sensitive to local history information, resulting in poor short-term prediction results. We not only design a novel encoder-decoder architecture to solve the problem of imbalanced performance of multi-step forecasting but also capture inter-spatial correlations dynamically. Therefore, we observe our CDGNet achieves state-of-the-art results on all tasks in Table \ref{main_res}.

\subsubsection{Results of the Encoder-Decoder Architecture}
In order to verify the effectiveness of our proposed encoder-decoder architecture for multi-step traffic forecasting, we use our architecture to replace the GMAN architecture for experiments. The experimental results of MAE on the METR-LA dataset are shown in Table \ref{arc}. As shown in Table \ref{arc}, compared with the original GMAN, using our proposed architecture achieves better results on all tasks and significantly reduces short-term prediction errors, because our architecture interacts with the encoder and is sensitive to local history information.
\subsubsection{Results on Weekdays and Weekends}
We further compare the forecasting performance of models in weekdays and weekends. Traffic flow on weekdays and weekends are widely divergent and jams are more likely to occur on weekdays, \emph{i.e.}, traffic conditions and the spatial dependence on the weekdays are more complicated. For example, people prefer to stay at home rather than go to office on weekends. Table \ref{week_res} shows the experimental results with weekdays and weekends on the METR-LA dataset. Our model achieves the best results for all periods of the weekdays and weekends. Weekends do not have obvious traffic trends like weekdays, so performance of weekends is worse than weekdays. Moreover, because our model can capture the inter-spatial dependence, it has a greater advantage on weekdays with more complex spatial correlations. Additionally, self-attention mechanism lacks inductive bias and requires a lot of data to learn dynamic spatial correlations; thus GMAN performs poorly on the weekends. Due to the few data on weekends, MTGNN is difficult to overfit and achieves better results. Of course, no matter how many samples, the learnable graph-based and dynamic graph-based methods are always better than static graph-based methods, such as DCRNN.
\begin{figure}[t]
    \centering
        \begin{subfigure}{0.32\linewidth}
        \imagebox{24mm}{\includegraphics[width=\linewidth,height=2.4cm]{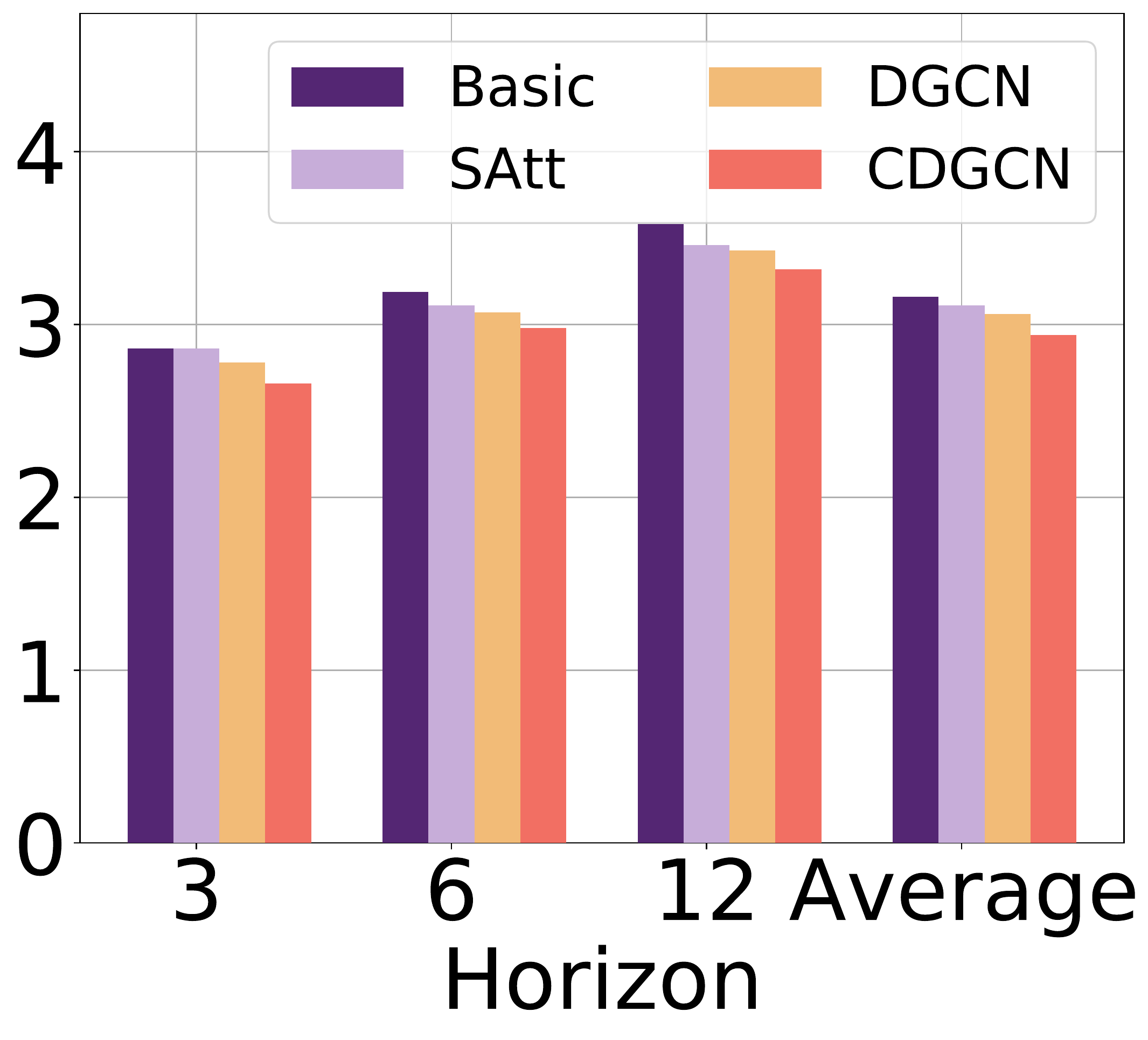}}
        \captionsetup{font=small,textfont=normalfont,labelfont=normalfont}
        \caption{MAE on METR-LA}
        \label{mae_metr}
      \end{subfigure}
      \hfill
      \begin{subfigure}{0.32\linewidth}
        \imagebox{24mm}{\includegraphics[width=\linewidth,height=2.4cm]{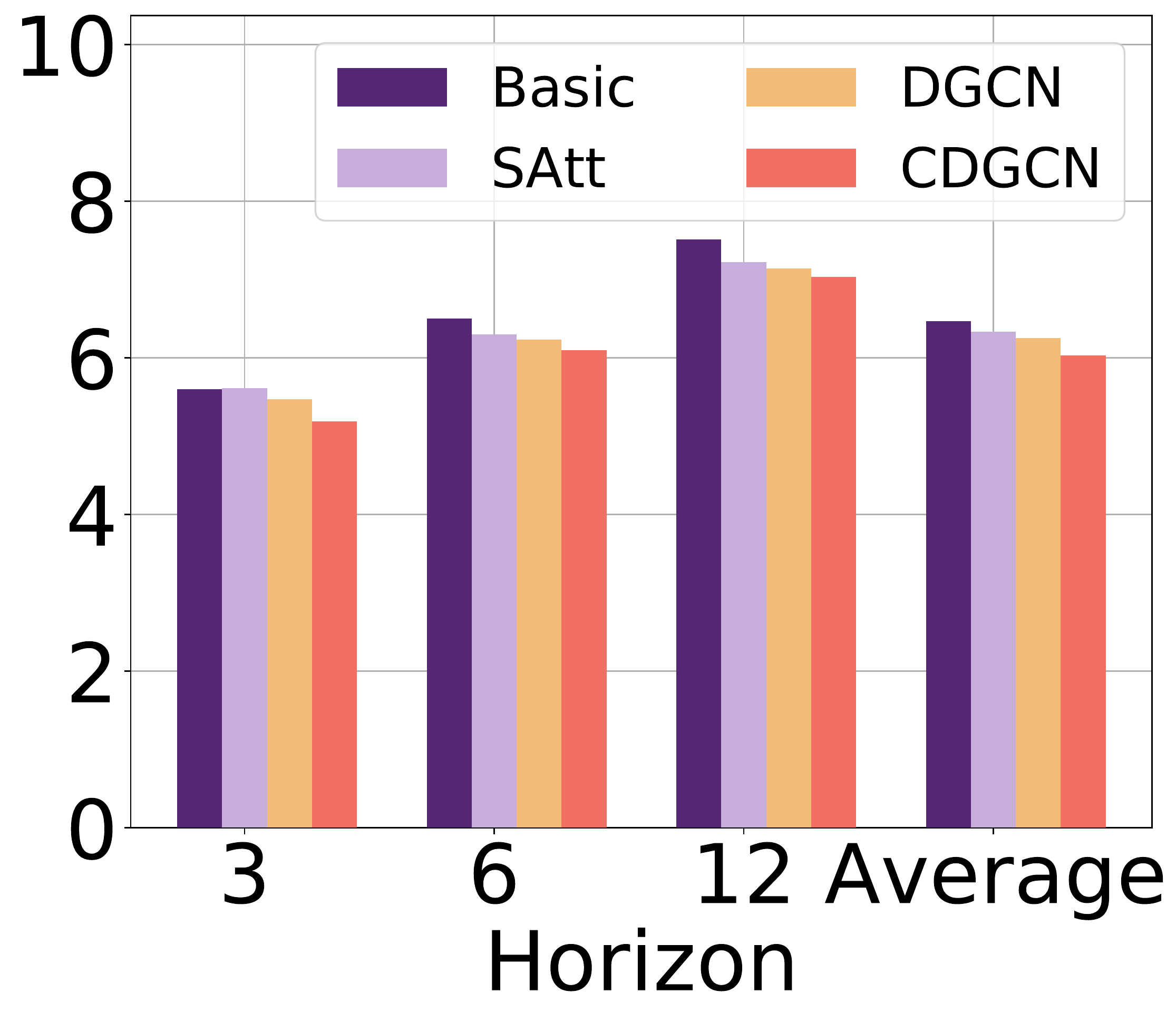}}
        \captionsetup{font=small,textfont=normalfont,labelfont=normalfont}
        \caption{RMSE on METR-LA}
        \label{rmse_metr}
      \end{subfigure}
      \hfill
      \begin{subfigure}{0.32\linewidth}
        \imagebox{24mm}{\includegraphics[width=\linewidth,height=2.4cm]{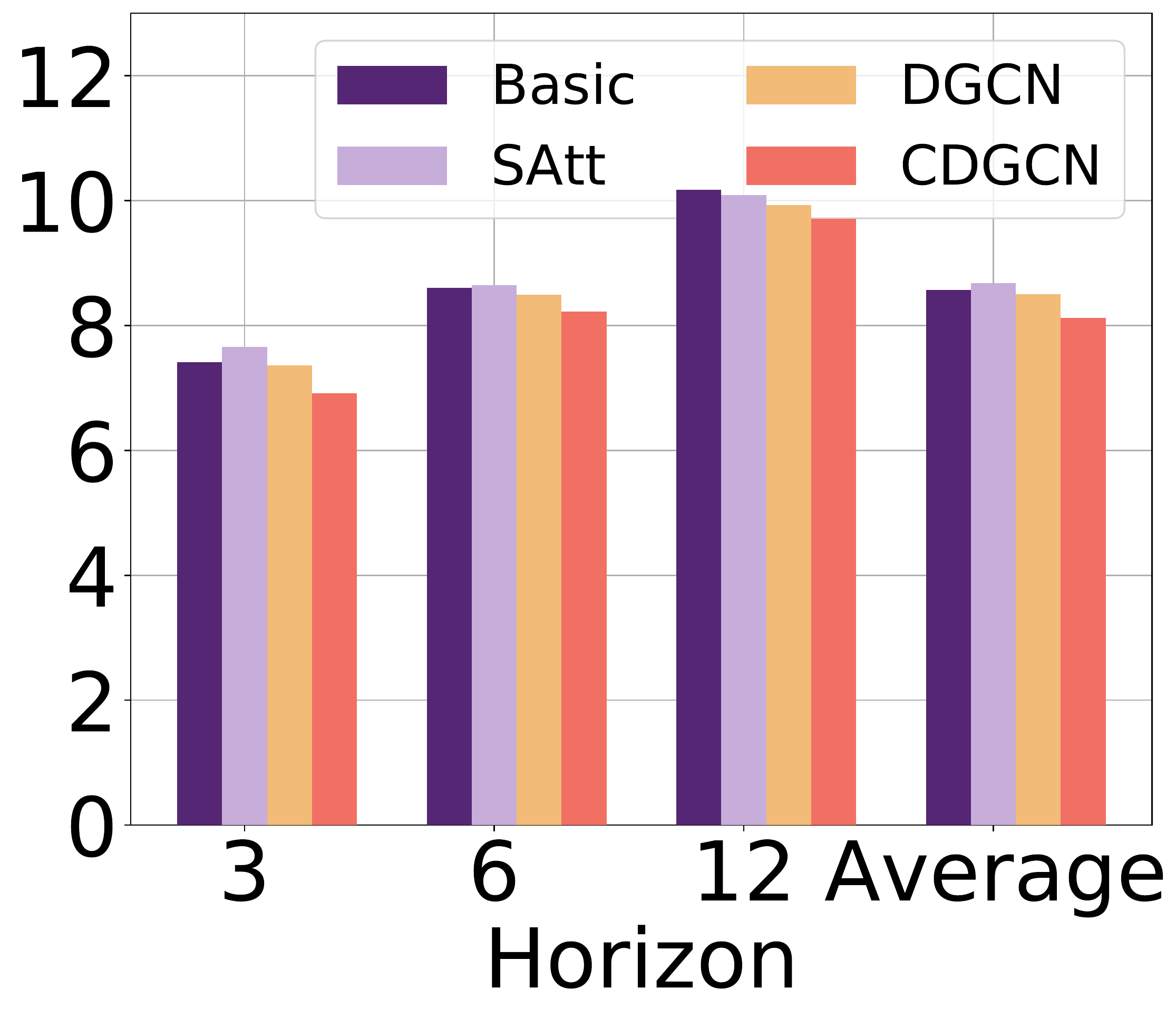}}
        \captionsetup{font=small,textfont=normalfont,labelfont=normalfont}
        \caption{MAPE on METR-LA}
        \label{mape_metr}
      \end{subfigure}
      
      \begin{subfigure}{0.32\linewidth}
        \imagebox{24mm}{\includegraphics[width=\linewidth,height=2.4cm]{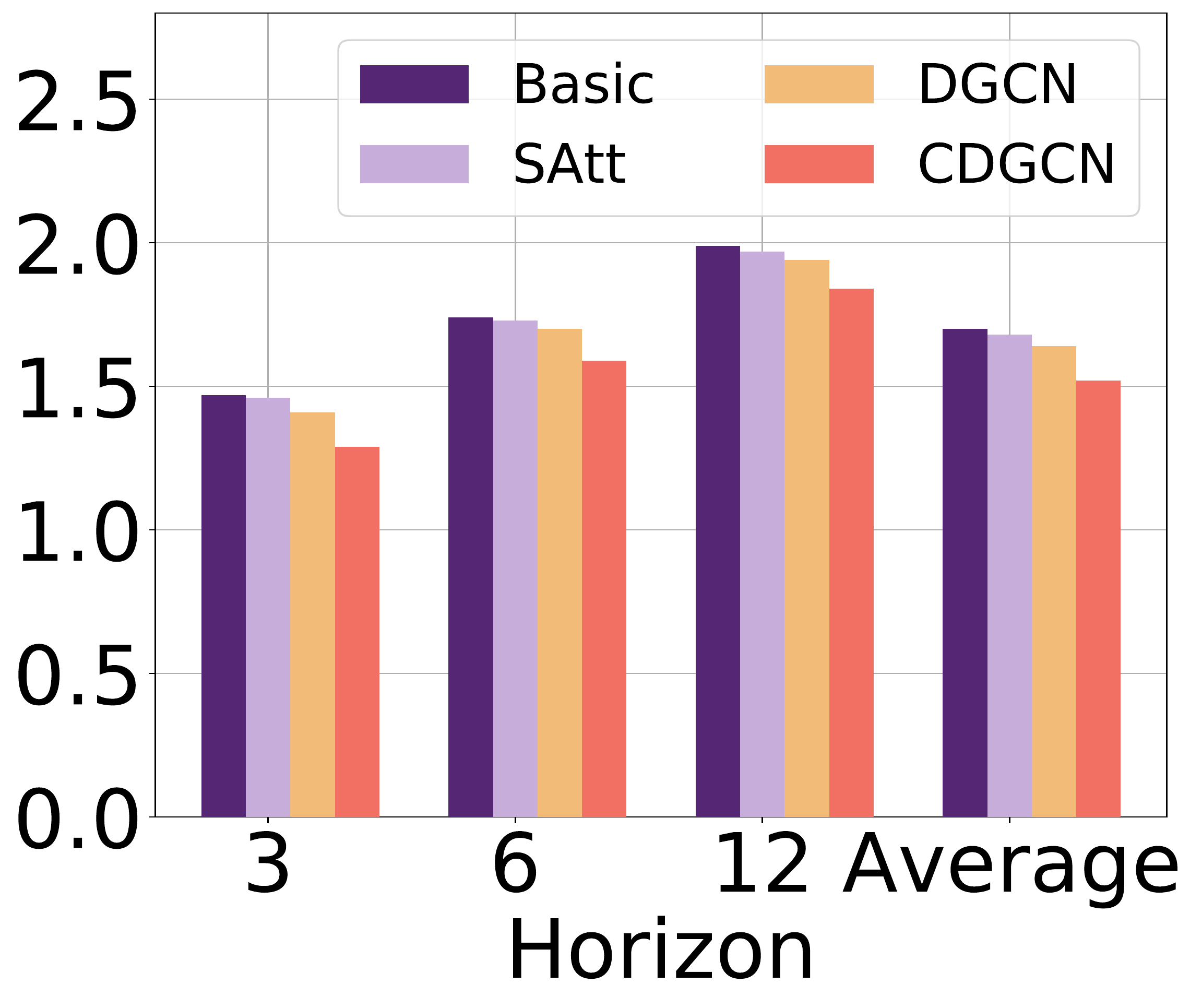}}
        \captionsetup{font=small,textfont=normalfont,labelfont=normalfont}
        \caption{MAE on PEMS-BAY}
        \label{mae_bay}
      \end{subfigure}
      \hfill
      \begin{subfigure}{0.32\linewidth}
        \imagebox{24mm}{\includegraphics[width=\linewidth,height=2.4cm]{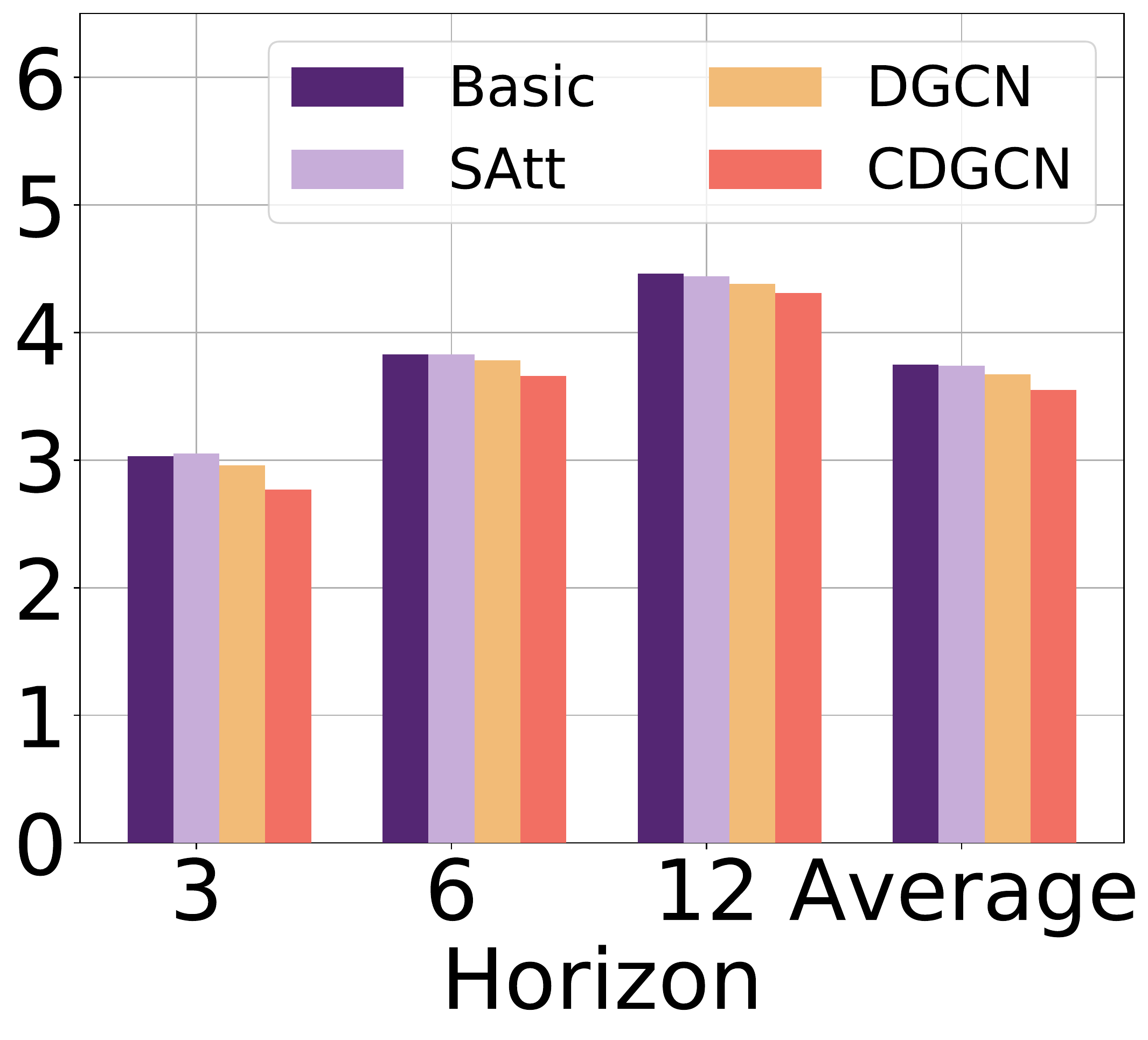}}
        \captionsetup{font=small,textfont=normalfont,labelfont=normalfont}
        \caption{RMSE on PEMS-BAY}
        \label{rmse_bay}
      \end{subfigure}
      \hfill
      \begin{subfigure}{0.32\linewidth}
        \imagebox{24mm}{\includegraphics[width=\linewidth,height=2.4cm]{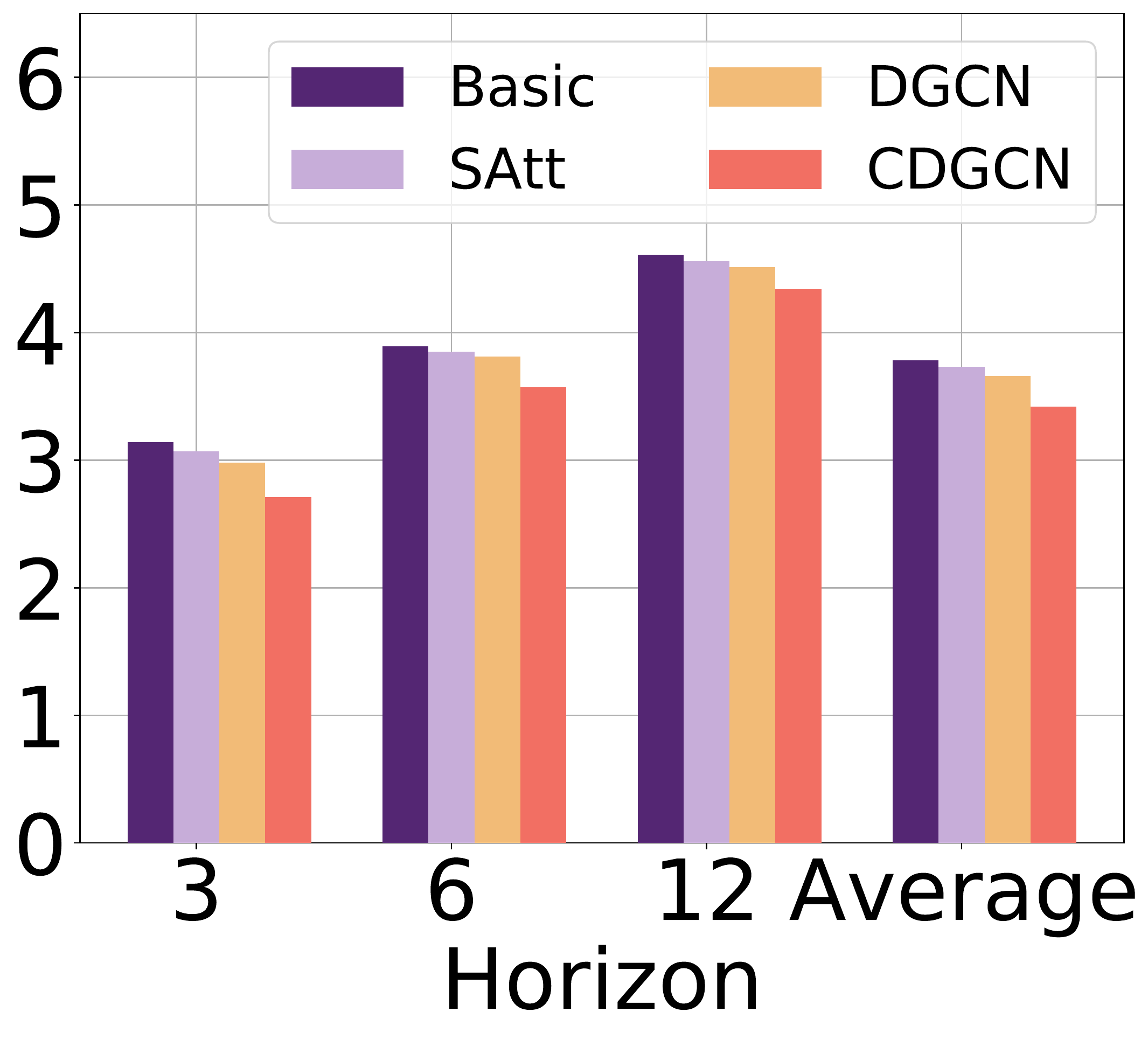}}
        \captionsetup{font=small,textfont=normalfont,labelfont=normalfont}
        \caption{MAPE on PEMS-BAY}
        \label{mape_bay}
      \end{subfigure}
      
      \begin{subfigure}{0.32\linewidth}
        \imagebox{24mm}{\includegraphics[width=\linewidth,height=2.4cm]{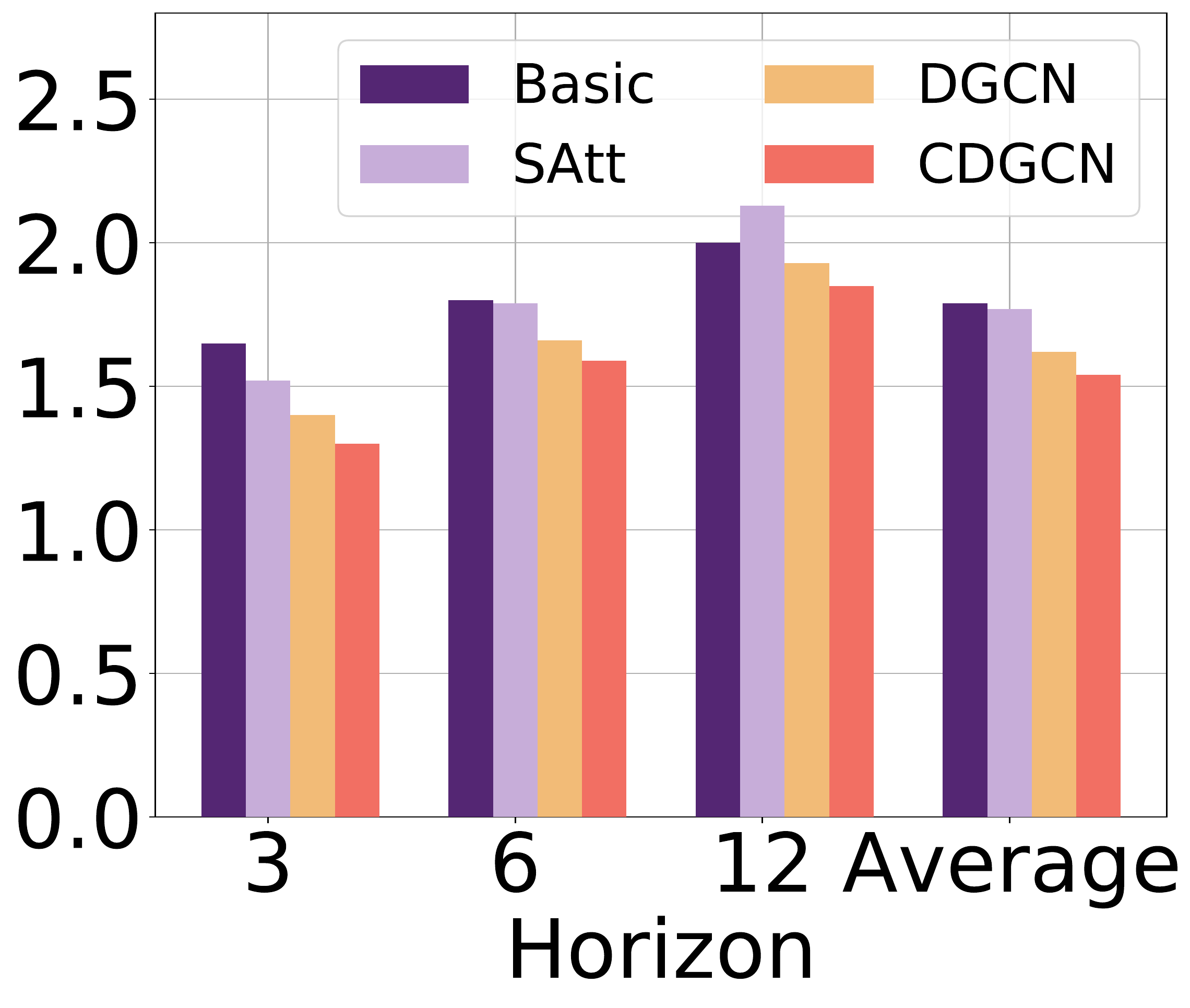}}
        \captionsetup{font=small,textfont=normalfont,labelfont=normalfont}
        \caption{MAE on PEMSD4}
        \label{mae_d4}
      \end{subfigure}
      \hfill
      \begin{subfigure}{0.32\linewidth}
        \imagebox{24mm}{\includegraphics[width=\linewidth,height=2.4cm]{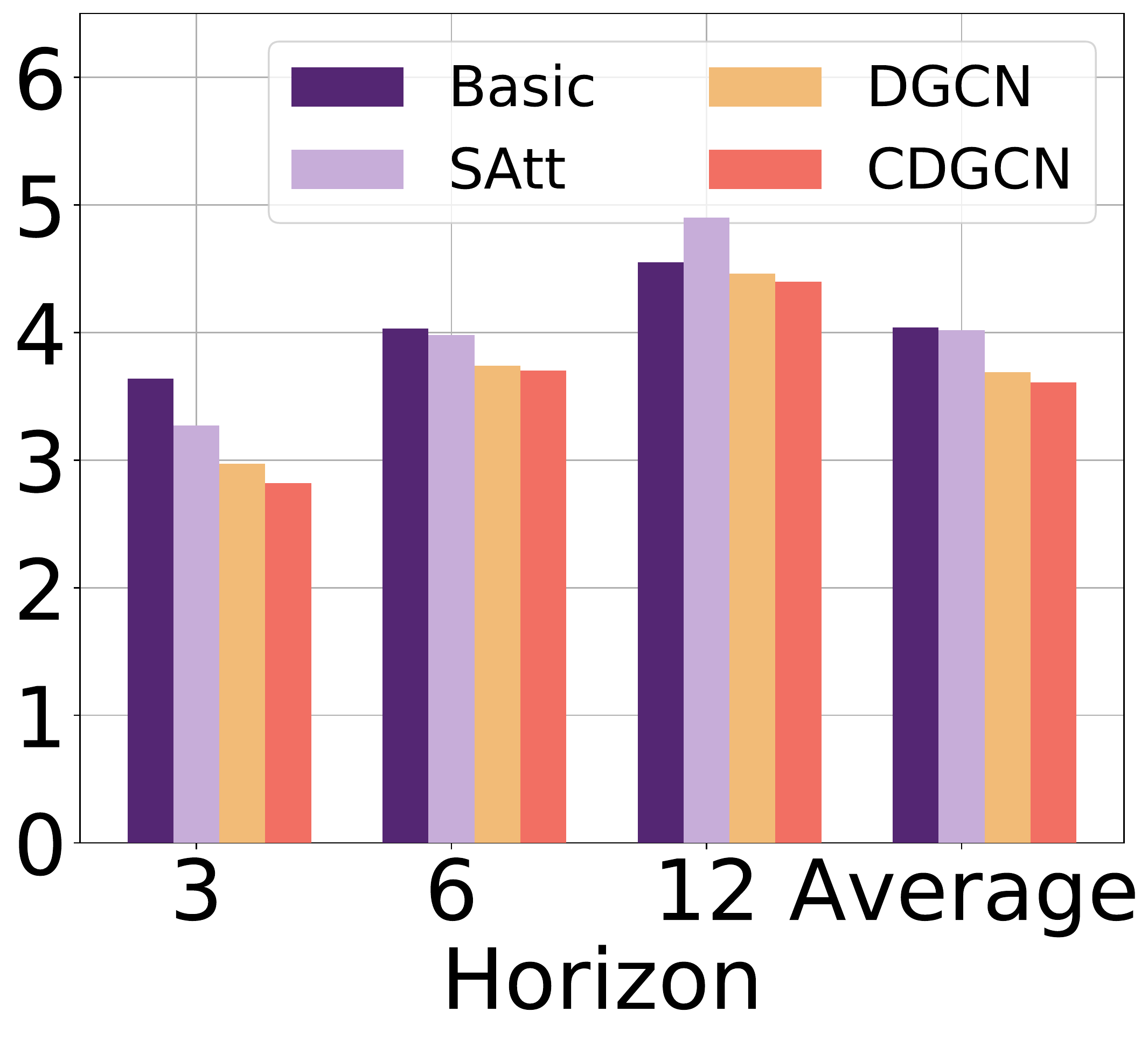}}
        \captionsetup{font=small,textfont=normalfont,labelfont=normalfont}
        \caption{RMSE on PEMSD4}
        \label{rmse_d4}
      \end{subfigure}
      \hfill
      \begin{subfigure}{0.32\linewidth}
        \imagebox{24mm}{\includegraphics[width=\linewidth,height=2.4cm]{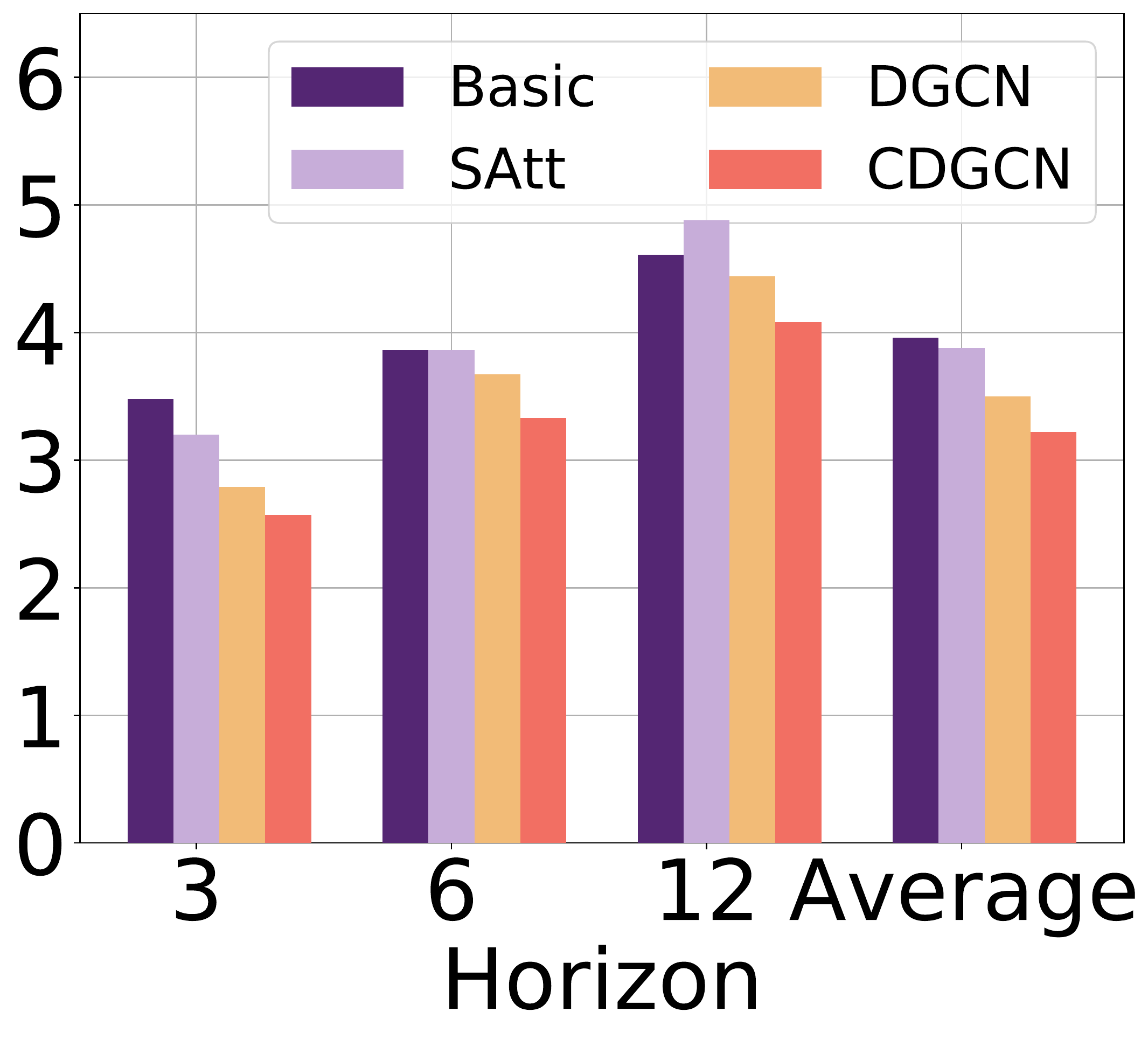}}
        \captionsetup{font=small,textfont=normalfont,labelfont=normalfont}
        \caption{MAPE on PEMSD4}
        \label{mape_d4}
      \end{subfigure}
      \vspace{-5pt}
      \caption{Ablation study.}
      \vspace{-12pt}
      \label{abl}
\end{figure}
\subsubsection{Ablation Study}
The two key points in our model are the dynamic calculation of the sparse adjacency matrix for each time slice in the form of graph convolution and the design of a cross-time mechanism to capture the inter-spatial dependence to avoid isolated modeling spatio-temporal dependencies, respectively. To verify the positive impact of each component in our model, we design three variants: Basic, SAtt, and DGCN:
\begin{itemize}[leftmargin=*]
    \item \textbf{Basic}: In this variant, we replace the CDGCN in the encoder-decoder architecture in CDGNet with the static distance graph-based vanilla GCN. Besides, we replace CDGCN-ED in the decoder with GCN-ED, which means that we utilize GCN between all future time slices and the last time slice of the encoder output.
    \item \textbf{SAtt}: In this variant, we replace the vanilla GCN in the encoder-decoder architecture in Basic with spatial self-attention. Besides, we replace GCN-ED in the decoder with SAtt-ED, which means that utilize spatial self-attention between all future time slices and the last time slice of the encoder output.
    \item \textbf{DGCN}: In this variant, we replace the spatial self-attention in the encoder-decoder architecture in SAtt with the dynamic graph convolution network in our paper. Besides, we replace SAtt-ED in the decoder with DGCN-ED, which means that utilize the dynamic graph convolution network between all future time slices and the last time slice of the encoder output.
\end{itemize}
We repeat each experiment 5 times with 10 epochs per repetition and report the average of 3 horizon, 6 horizon, 12 horizon, and average results of MAE, RMSE, MAPE on three datasets in Figure \ref{abl}. Compared to Basic, SAtt achieves better performance, indicating that the dynamic graph based on the temporal feature reflects conditions of the road network better than the static graph based on distance. Compared to SAtt, DGCN performs better demonstrates importance of the sparse graph in traffic forecasting. The outperformance of CDGCN over DGCN indicates the importance of capturing the inter-spatial dependence across time.

\begin{figure}[t]
    \centering
        \begin{subfigure}{0.33\linewidth}
        \includegraphics[width=\linewidth]{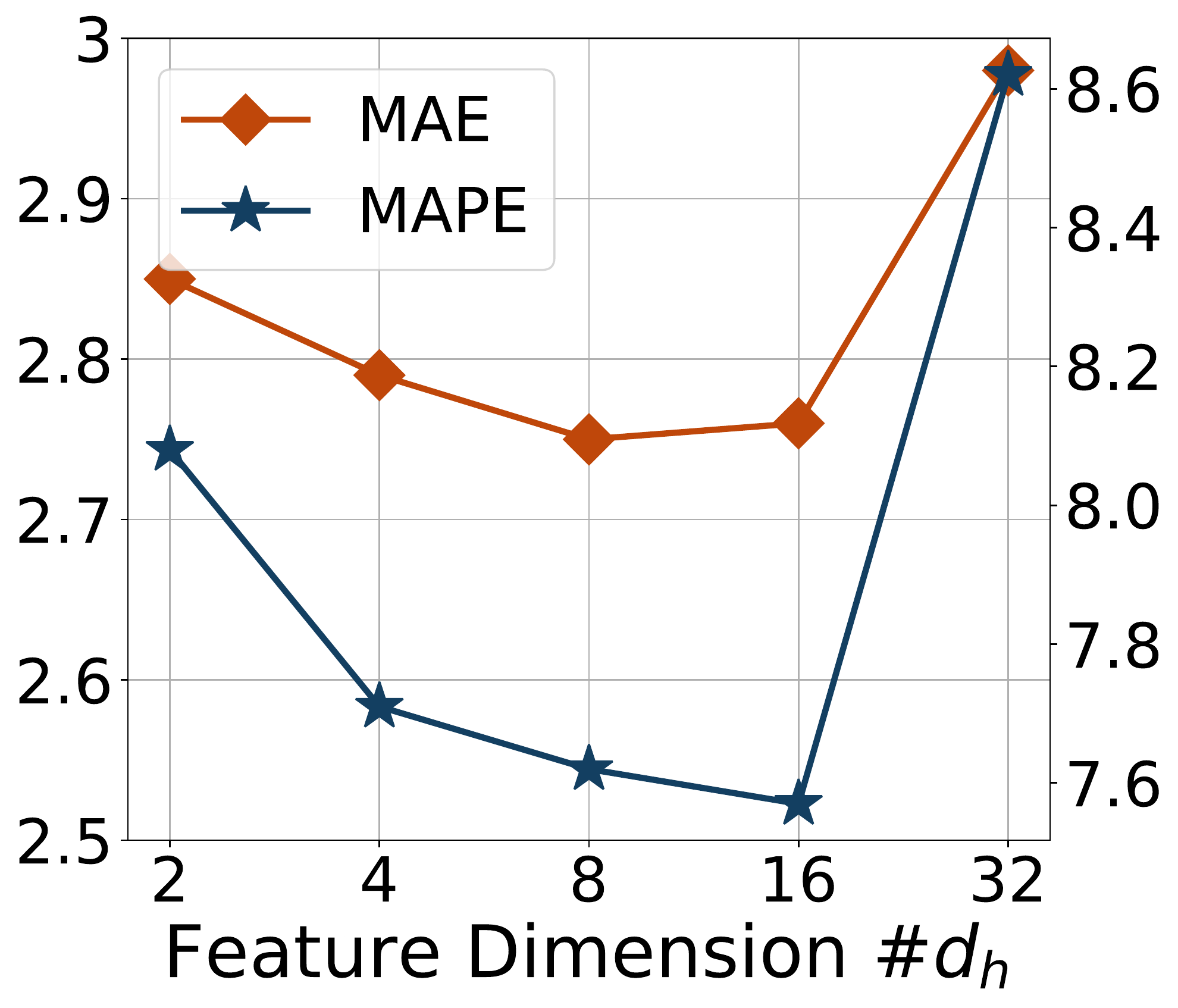}
        \captionsetup{font=small}
        \caption{}
        \label{d}
      \end{subfigure}%
      \hfill
      \begin{subfigure}{0.33\linewidth}
        \includegraphics[width=\linewidth]{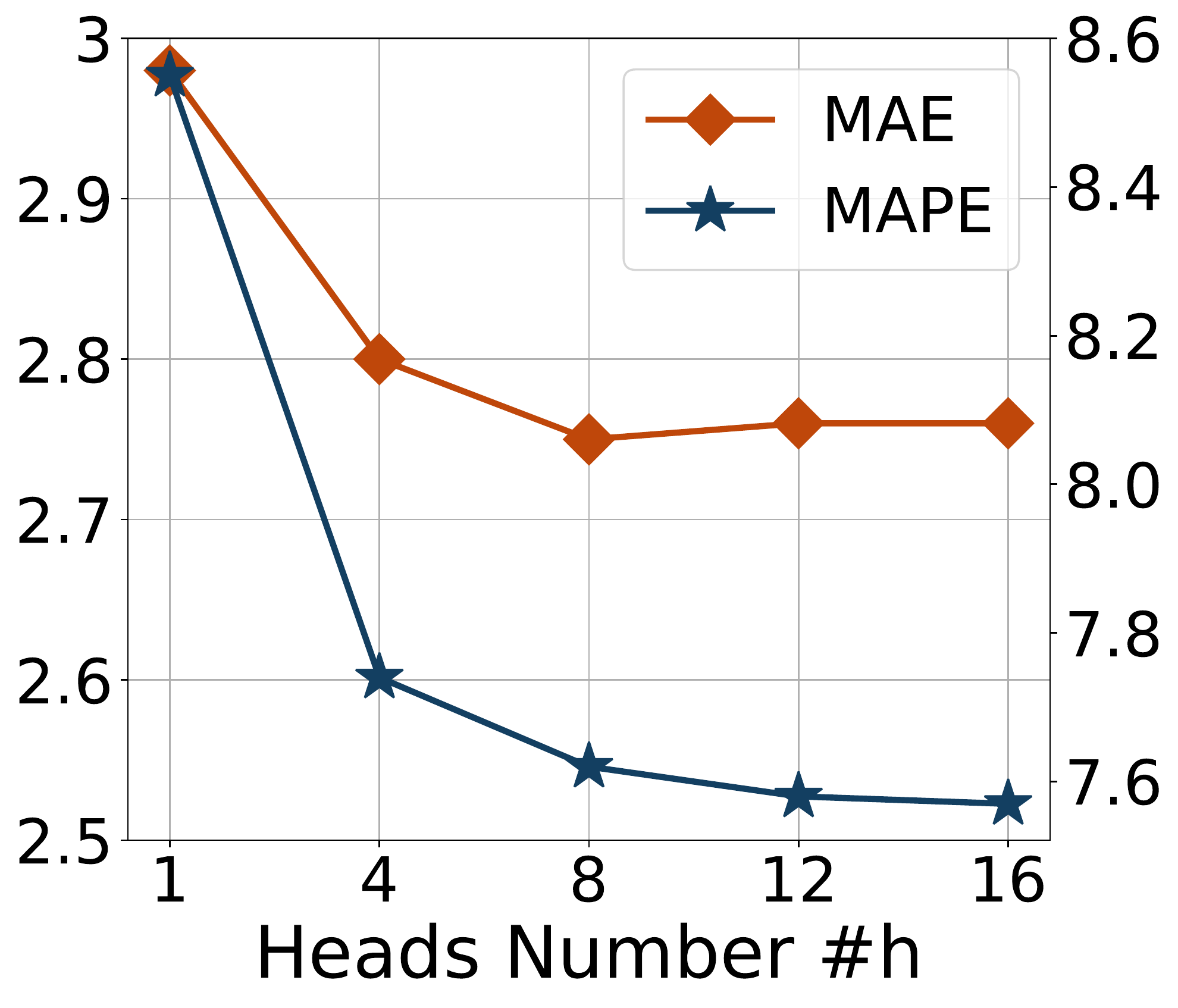}
        \captionsetup{font=small}
        \caption{}
        \label{h}
      \end{subfigure}
      \hfill
      \begin{subfigure}{0.33\linewidth}
        \includegraphics[width=\linewidth]{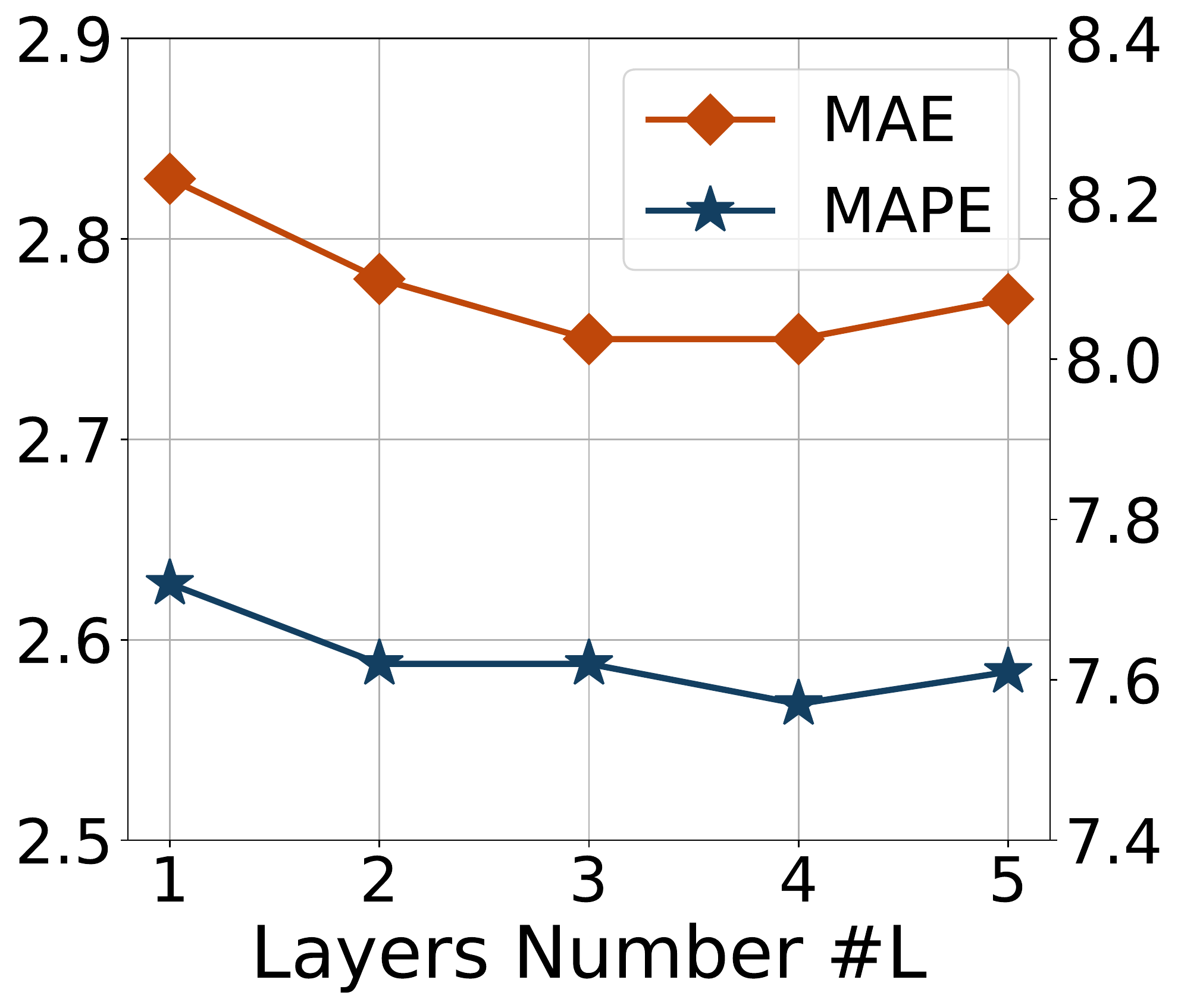}
        \captionsetup{font=small}
        \caption{}
        \label{L}
      \end{subfigure}
      \caption{Hyper-parameter study on the METR-LA dataset.}
      \vspace{-10pt}
      \label{param}
\end{figure}
\begin{figure}[t]
    \centering
    \begin{subfigure}{0.48\linewidth}
        \includegraphics[width=\linewidth]{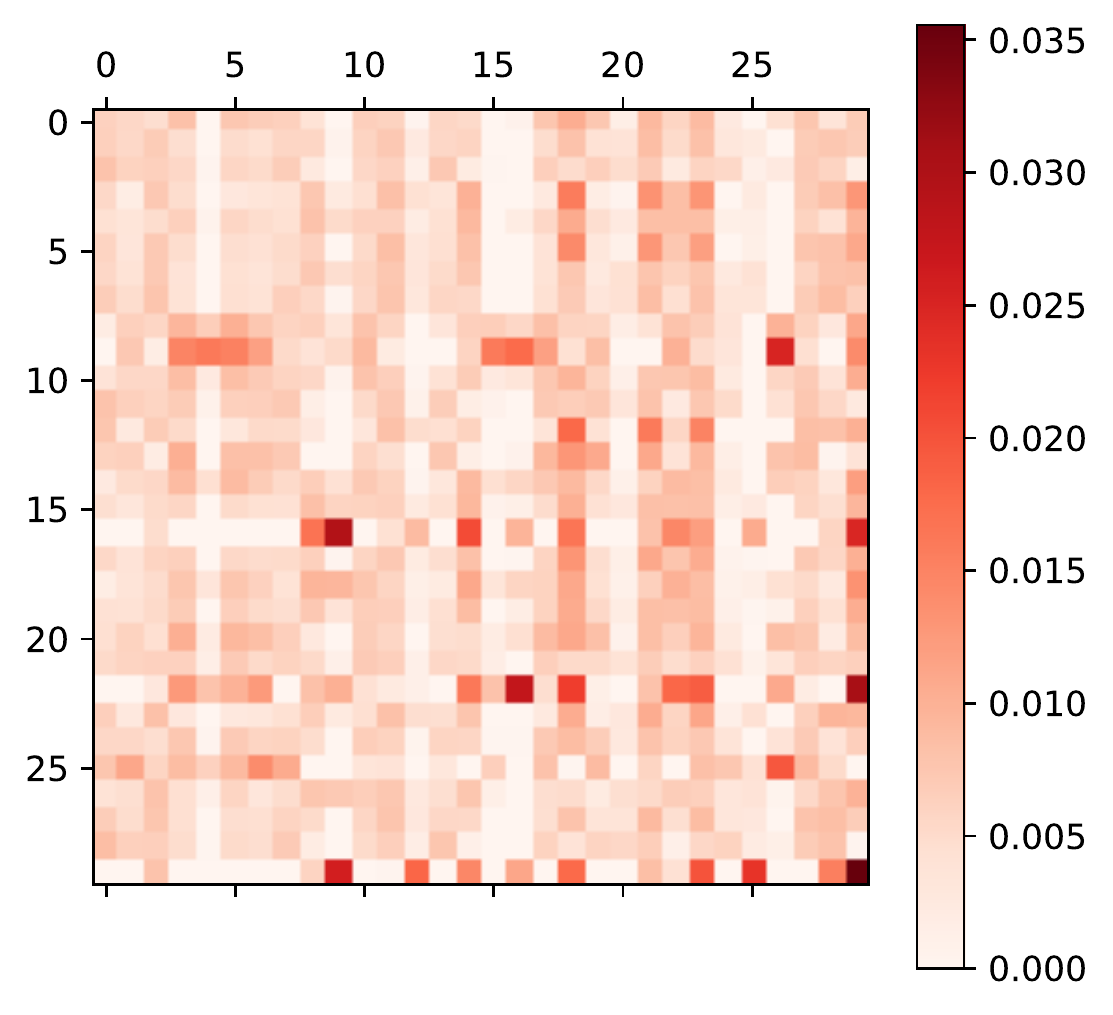}
        \captionsetup{font=small}
        \caption{Adjacency matrix for the first 29 sensors in the morning on 6/30/2012.}
        \label{vis_morning}
      \end{subfigure}
      \hfill
      \begin{subfigure}{0.48\linewidth}
        \includegraphics[width=\linewidth]{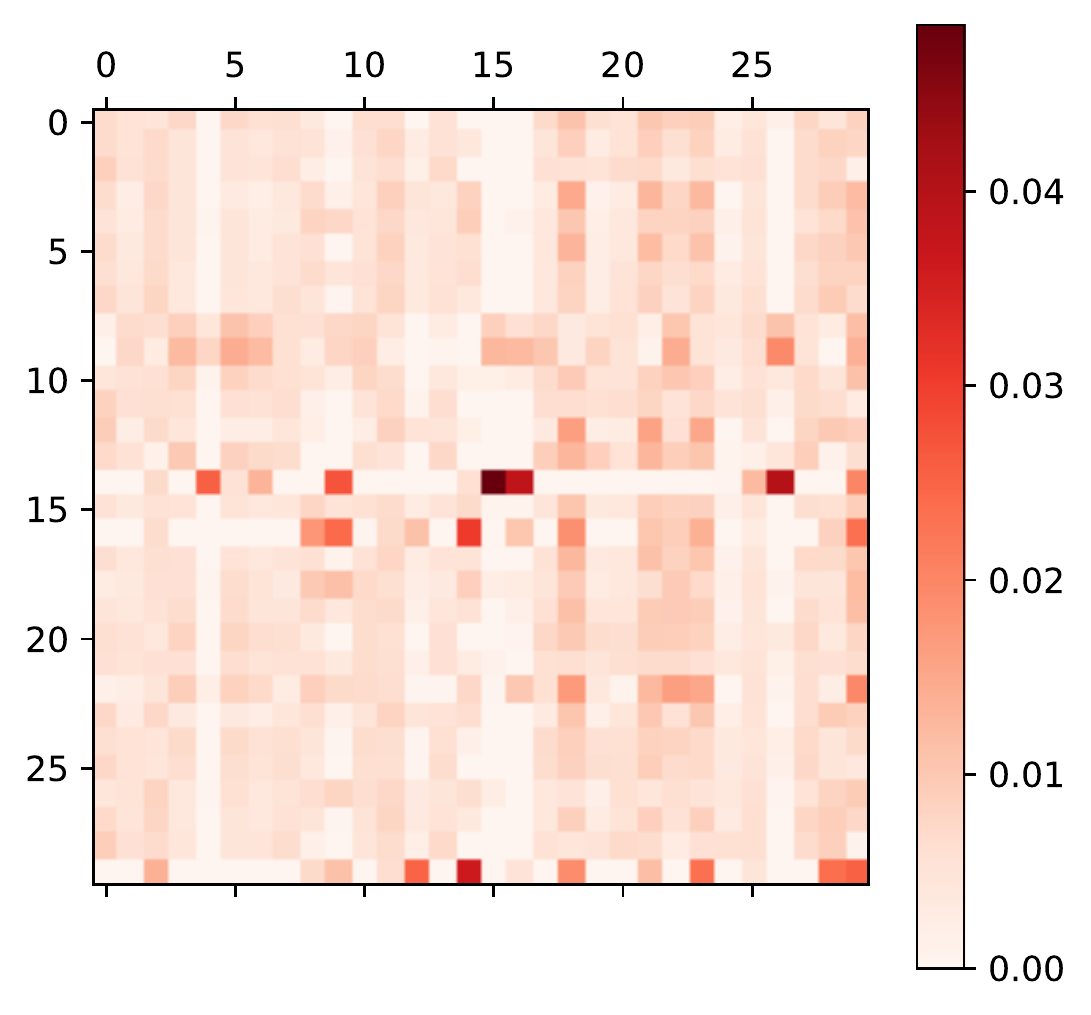}
        \captionsetup{font=small}
        \caption{Adjacency matrix for the first 29 sensors in the evening on 6/30/2012.}
        \label{vis_night}
      \end{subfigure}
      
      \begin{subfigure}{1.0\linewidth}
        \includegraphics[width=\linewidth]{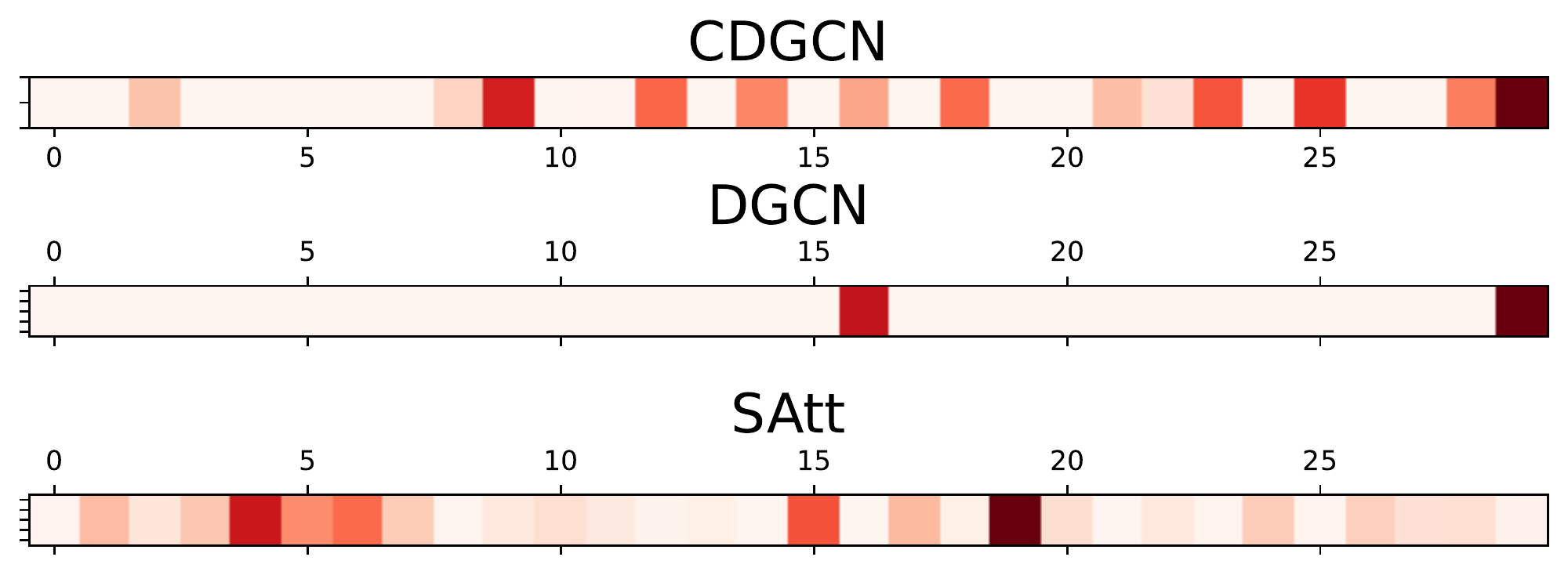}
        \captionsetup{font=small}
        \caption{The enlarged view of correlations of the $29$-th sensor under different methods in the morning.}
        \label{vis_29}
      \end{subfigure}
      
      \begin{subfigure}{1.0\linewidth}
        \includegraphics[width=\linewidth]{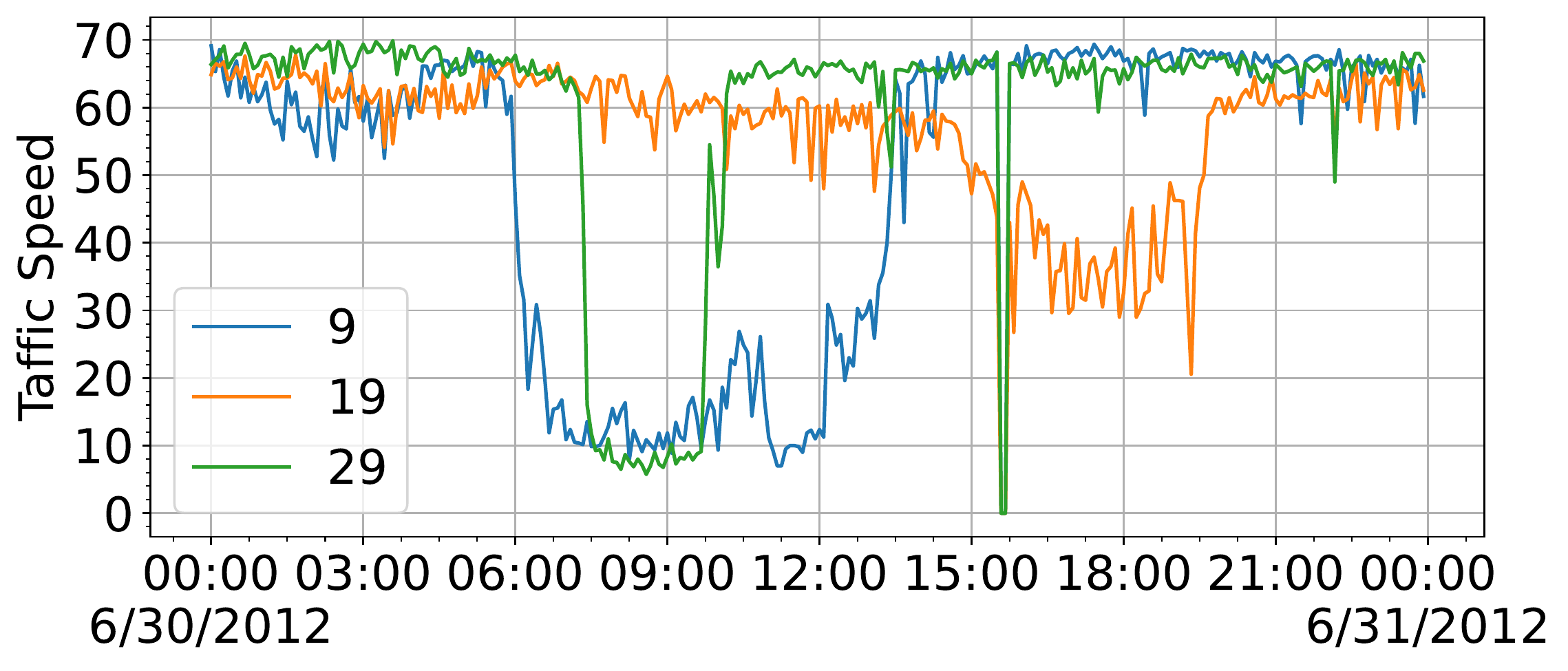}
        \captionsetup{font=small}
        \caption{Traffic speed of sensor 9, 19, and 29.}
        \label{vis_speed}
      \end{subfigure}
      
      \vspace{-6pt}
      \caption{Visualization on the METR-LA dataset.}
      \vspace{-20pt}
      \label{vis}
\end{figure}
\subsubsection{Hyper-Parameter Study}
We conduct a hyper-parameter study on three core hyper-parameters which affect the model complexity and list these hyper-parameters: Number of layers in CDGNet, we conduct a grid search over $L=\{1,2,3,4,5\}$. Number of head in CDGCN, we conduct a grid search over $h=\{1,4,8,12,16\}$. Number of feature dimension of single head in CDGCN, we conduct a grid search over $d_h=\{2,4,8,16,32\}$. We repeat each experiment 5 times with 10 epochs each time and report the average of MAE and MAPE on the validation set of METR-LA. The default settings are $L=3$, $h=8$, and $d_h=8$. As shown in Figure \ref{d}, appropriately increasing the feature dimension can improve performance of our model, but too large feature dimension leads to over-fitting. Results in Figure \ref{h} indicate that when the number of heads in CDGCN is large, increasing the number of heads is not cost-effective in terms of consumption and performance. Compared with the number of heads and feature dimensions, our model is not sensitive to the number of layers as the results in Figure \ref{L}, but a moderate increase in the number of layers can also improve performance.
\subsubsection{Visualization}
In Figure \ref{vis_morning} and \ref{vis_night}, we visualize the adjacency matrix for the first 29 sensors in the morning and evening on 6/30/2012 in METR-LA, and we observe that they are different and correlations are denser in the morning, because of the dynamic spatial dependence. We further enlarge spatial correlations of the $29$-th sensor under different methods in Figure \ref{vis_29}. SAtt does not learn the self-correlation and considers the $19$-th sensor is important, but temporal patterns of $19$-th sensor and $29$-th sensor are completely inconsistent in Figure \ref{vis_speed}. Besides, as shown in Figure \ref{vis_speed}, the rush hour in the morning at the location of $9$-th sensor is earlier than $29$-th sensor, \emph{i.e.}, the traffic flow pass to $29$-th sensor after a period of propagation. However, SAtt and DGCN fail to capture the correlation between $29$-th sensor and $9$-th sensor, \emph{i.e.}, they fail to capture the cross-time spatial dependence with hysteresis.
\section{conclusion}
In this paper, we propose a novel CDGNet to forecast traffic speed. We design a novel cross-time dynamic graph-based GCN for each time slice to capture intra-spatial dependence and inter-spatial dependence. Besides, we use a gating mechanism to sparse the cross-time dynamic graph. To make better use of CDGCN, we design a novel encoder-decoder architecture to balance the performance of long short-term forecasting. We evaluate CDGNet on three real-world traffic datasets and it achieves state-of-the-art performance on all tasks compared with baselines. In the future, we will apply our model to other multivariate time series forecasting tasks.

\bibliographystyle{ACM-Reference-Format}
\bibliography{cdgcn}
\end{document}